\documentclass{article}

\PassOptionsToPackage{numbers, compress}{natbib}

\usepackage[preprint]{neurips_2025}

\usepackage[utf8]{inputenc} 
\usepackage[T1]{fontenc}    
\usepackage{hyperref}       
\usepackage{url}            
\usepackage{booktabs}       
\usepackage{amsfonts}       
\usepackage{nicefrac}       
\usepackage{microtype}      
\usepackage{xcolor}         
\usepackage{graphicx}       
\usepackage{amsmath}
\usepackage{arydshln}

\NewDocumentCommand{\zihao}{ mO{} }{\textcolor{red}{\textsuperscript{\textit{zihao}}\textsf{\textbf{\small[#1]}}}}

\title{Localizing Knowledge in Diffusion Transformers}

\author{Arman Zarei\textsuperscript{1}\thanks{Correspondence to: azarei@umd.edu} \hspace{2pt},\hspace{2pt} Samyadeep Basu\textsuperscript{1,3},\hspace{2pt} Keivan Rezaei\textsuperscript{1},\hspace{2pt} Zihao Lin\textsuperscript{2},\hspace{2pt} Sayan Nag\textsuperscript{3},\hspace{2pt} Soheil Feizi\textsuperscript{1}\\
  \textsuperscript{1}University of Maryland \hspace{14pt} \textsuperscript{2}University of California, Davis \hspace{14pt} \textsuperscript{3}Adobe
}

\begin{document}

\maketitle

\begin{abstract}

Understanding how knowledge is distributed across the layers of generative models is crucial for improving interpretability, controllability, and adaptation. While prior work has explored knowledge localization in UNet-based architectures, Diffusion Transformer (DiT)-based models remain underexplored in this context. In this paper, we propose a model- and knowledge-agnostic method to localize where specific types of knowledge are encoded within the DiT blocks. We evaluate our method on state-of-the-art DiT-based models, including PixArt-$\alpha$, FLUX, and SANA, across six diverse knowledge categories. We show that the identified blocks are both interpretable and causally linked to the expression of knowledge in generated outputs.
Building on these insights, we apply our localization framework to two key applications: {\it model personalization} and {\it knowledge unlearning}. In both settings, our localized fine-tuning approach enables efficient and targeted updates, reducing computational cost, improving task-specific performance, and better preserving general model behavior with minimal interference to unrelated or surrounding content.
Overall, our findings offer new insights into the internal structure of DiTs and introduce a practical pathway for more interpretable, efficient, and controllable model editing. \footnote[1]{Project page is available at: \href{https://armanzarei.github.io/Localizing-Knowledge-in-DiTs}{https://armanzarei.github.io/Localizing-Knowledge-in-DiTs}}
\end{abstract}

\section{Introduction}
\label{sec:introduction}

Diffusion and flow models \cite{ho2020denoising, rombach2022high, nichol2021improved, ho2022classifier, flux2024, esser2024scaling, lipman2022flow} have rapidly become the leading paradigm for a wide range of generative tasks, particularly text-to-image (T2I) synthesis. With access to such powerful pretrained models, it is crucial to explore their potential for applications beyond mere generation. A growing body of work \cite{hertz2022prompt, tumanyan2023plug, kumari2023ablating, zarei2024understanding} has focused on localizing different types of knowledge and capabilities within these models, enabling more targeted usage. For example, \cite{hertz2022prompt} showed that cross-attention layers are key to incorporating prompt compositional information, while \cite{tumanyan2023plug, liu2024towards} demonstrated that structural information is often concentrated in the self-attention modules of UNet-based architectures. These insights have been applied to tasks such as image editing and structure-preserving generation \cite{hertz2022prompt, cao2023masactrl, liu2024towards}.

Localizing where specific knowledge resides within models is essential for interpretability, targeted interventions, and understanding model behavior. In generative models, it plays a critical role in applications such as model unlearning and personalization. Several works \cite{kumari2023ablating, schramowski2023safe, gandikota2024unified, yoon2024safree} have shown that generative models often memorize unsafe or unwanted content (e.g., copyrighted or NSFW content) and proposed methods to selectively erase such concepts. In model personalization, the goal is to generate novel renditions of a subject using only a few reference photos across diverse scenes and poses. In both cases, localizing knowledge within the model is crucial for enabling {\it targeted interventions} that make fine-tuning more efficient and effective, while better preserving the model's prior capabilities and overall generation quality.

Recent advances in text-to-image generation have marked a notable evolution, transitioning from UNet-based architectures \cite{ronneberger2015u} to Transformer-based models \cite{vaswani2017attention}, particularly the Diffusion Transformer (DiT) \cite{peebles2023scalable}. DiT architectures, with their purely attention-based structure, have achieved state-of-the-art generation quality compared to UNet counterparts. While extensive research has explored interpretability and localization in UNet-based architectures, DiT-based models have received comparatively little attention in this regard despite their recent emergence and strong performance. 

\begin{figure}[t]
    \centering
    \includegraphics[width=\linewidth]{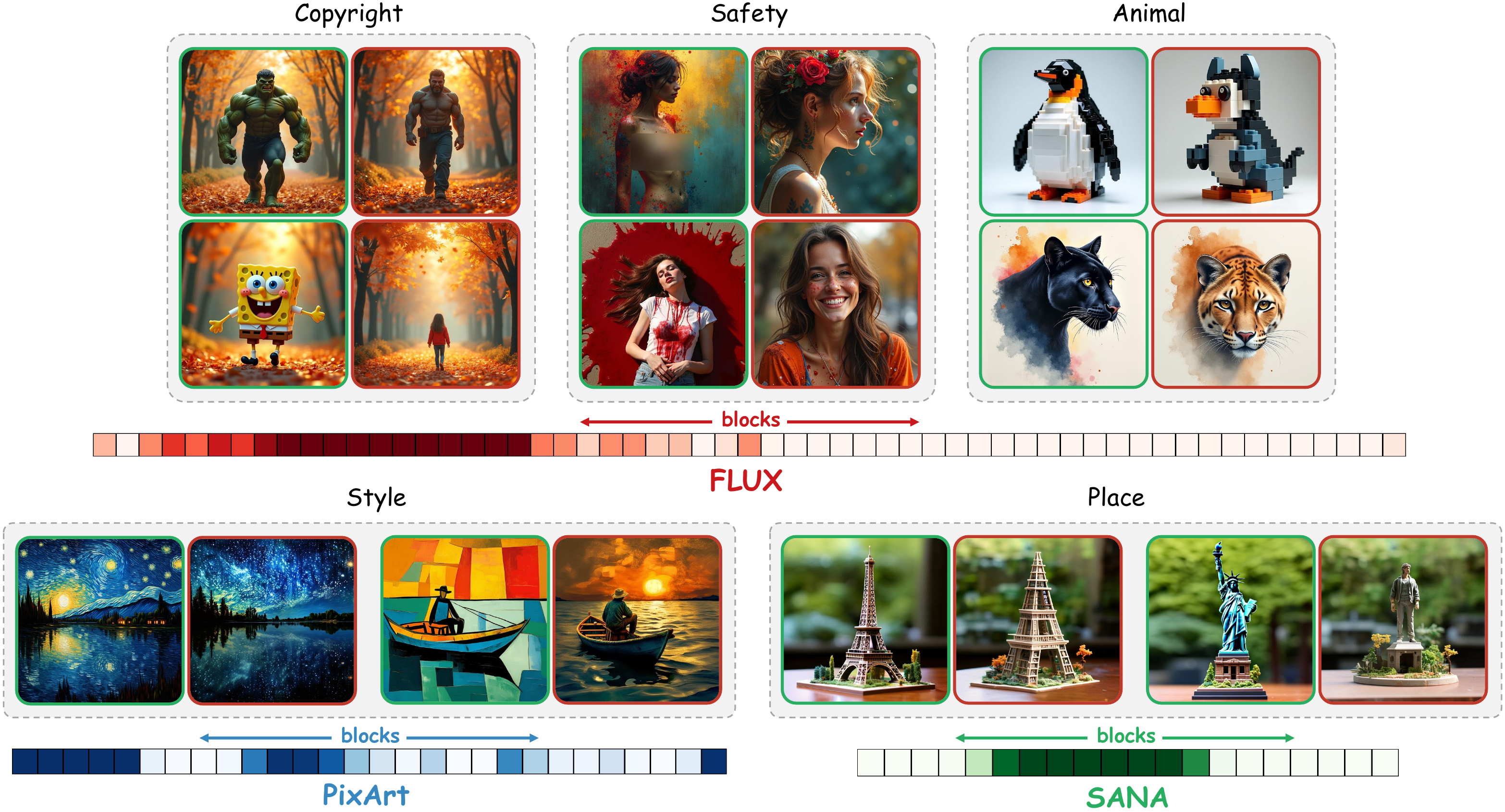}
    \caption{\textbf{Localization across various DiT models and knowledge categories.} For each model, heatmaps indicate the frequency of each block being selected as a dominant carrier of different target knowledge. \textcolor[HTML]{27ae60}{Green}-bordered images are standard generations, while \textcolor[HTML]{C0392B}{red}-bordered images result from withholding knowledge-specific information in the localized blocks. Our method successfully localizes diverse knowledge types, with variation in localization patterns across models.}
    \label{fig:localization_categories_in_different_models}
\end{figure}

In this paper, we thoroughly investigate the localization of different types of knowledge within the blocks of diffusion transformers across a range of state-of-the-art models, including FLUX \cite{flux2024}, SANA \cite{xie2024sana}, and PixArt-$\alpha$ \cite{chen2023pixart}. We introduce a model-agnostic and knowledge-agnostic method that provides a strong and reliable signal for identifying the blocks most responsible for generating specific types of knowledge. Our approach demonstrates strong performance and robustness across all evaluated models and a diverse set of knowledge categories, such as copyrighted content, NSFW material, and artistic styles (Figure~\ref{fig:localization_categories_in_different_models}). While the distribution of localized blocks varies from model to model, our method consistently identifies the key regions responsible for encoding each knowledge.

\begin{figure}[t]
    \centering
    \includegraphics[width=\linewidth]{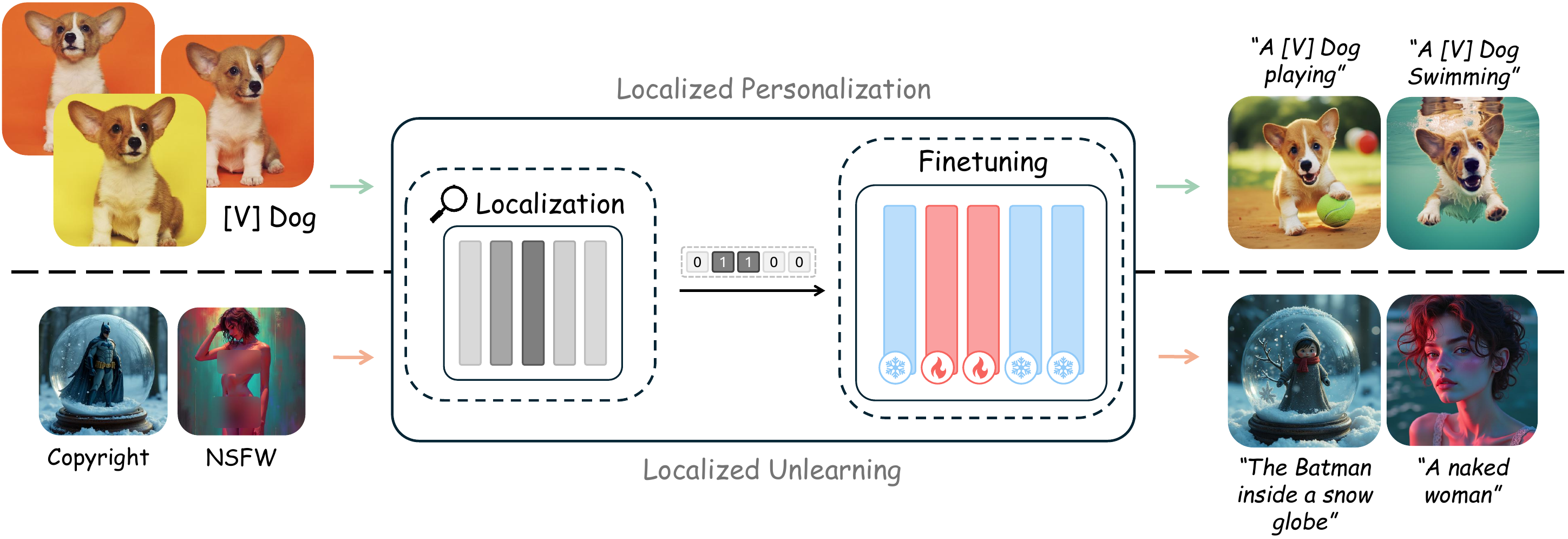}
    \caption{\textbf{Targeted fine-tuning via knowledge localization.} Given a concept to personalize or remove, our method first identifies the most relevant blocks via knowledge localization and restricts fine-tuning to those blocks. This enables efficient adaptation (top) and targeted suppression (bottom) with minimal impact on surrounding content, while better preserving the model’s prior performance.}
    \label{fig:localization_applications}
\end{figure}

Building upon our knowledge localization technique, we further propose practical applications of our method for model personalization and unlearning in DiTs (Figure \ref{fig:localization_applications}). Whether the objective is to inject new knowledge or remove undesired content, our approach first localizes the relevant information within the blocks of the DiT and then enables targeted interventions to modify it. Through extensive experiments, we demonstrate that our method outperforms baseline approaches that modify all blocks, achieving superior preservation of generation quality and consistency on unrelated and surrounding prompts. Notably, in the personalization setting, our method achieves improved task-specific performance compared to full-model fine-tuning, while also minimizing interference with surrounding knowledge. Additionally, our method is more efficient, offering faster training and lower memory usage compared to these baselines.

In summary, our contributions are: 
(1) We are the first to explore knowledge localization in DiTs by introducing a large-scale probing dataset covering diverse categories, and by proposing an automatic, model- and knowledge-agnostic method for identifying where such information resides within the model’s blocks.
(2) We conduct extensive evaluations across multiple DiT architectures and diverse knowledge types to validate the generality and robustness of our approach. 
(3) Building on this localization, we demonstrate practical applications for efficient model personalization and unlearning. Our method enables targeted fine-tuning that is faster and more memory-efficient, while also achieving superior preservation of generation quality and consistency on unrelated or surrounding prompts.

\section{Related Works}
\subsection{Diffusion and Flow Models}

Diffusion models are a class of generative models based on stochastic differential equations (SDEs), where noise is progressively added to data through a stochastic forward process, ultimately transforming the data distribution into a standard Gaussian. A corresponding reverse process is then learned to reconstruct the original data from noise. Flow matching methods, closely related to diffusion models, instead define a deterministic mapping from noise to data using an ordinary differential equation (ODE). These methods learn a time-dependent vector field $v_\theta(x,t)$ that is trained to approximate a target field $v_t(x)$, which directs samples from the noise distribution toward the data distribution, by minimizing a flow matching loss. See Appendix~\ref{app:related_works_diffusion_and_flow} for further details.

\subsection{Interpretability of Diffusion Transformers}
The internal mechanisms of text-to-image diffusion models have been primarily explored in the context of UNet-based models~\citep{basu2024localizing, basu2024on, zarei2024understanding}. These studies reveal that knowledge of various visual concepts—such as artistic style—is either localized or distributed across a small subset of layers within the UNet architecture. Beyond offering interpretability, these localization insights have been leveraged to address practical challenges, including the removal of copyrighted content, without the need for full model retraining. With the recent shift toward transformer-based models such as Flux~\citep{flux2024} and PixArt-$\alpha$~\citep{chen2023pixart}, understanding how and where concepts are encoded in these new models has become an emerging area of interest. 
Recent work has begun to uncover the interpretability of diffusion transformers. For instance, \citep{helbling2025conceptattentiondiffusiontransformerslearn} show that attention maps in models like Flux can act as high-quality saliency maps, while \citep{avrahami2025stableflowvitallayers} identify a subset of critical layers that are particularly effective for downstream tasks such as image editing. 
However, it remains unclear how knowledge of visual concepts—such as copyrighted objects, artistic styles, or safety-related content—is represented and localized within diffusion transformers. Gaining such insights could enable targeted removal of undesirable content and enhance model personalization for various downstream applications.  

\section{Localizing Knowledge in Diffusion Transformers}

\subsection{Probe Dataset Description}

\newcommand{\datasetname}{\mathcal{L}\textsc{oc}\mathcal{K}}

To systematically evaluate the generalizability and robustness of our localization method, we first introduce a new dataset called $\datasetname$ (\underline{$\mathcal{L}$oc}alization of \underline{$\mathcal{K}$}nowledge) designed around six distinct categories of knowledge and concepts: artistic styles (e.g., \textit{“style of Van Gogh”}), celebrities (e.g., \textit{“Albert Einstein”}), sensitive or safety-related content (e.g., \textit{“a naked woman”, “a dead body covered in blood”}), copyrighted characters (e.g., \textit{“the Batman”}), famous landmarks (e.g., \textit{“the Eiffel Tower”}), and animals (e.g., \textit{“a black panther”}). These categories are selected to cover a diverse range of visual and semantic information, while also being representative of key use cases in model unlearning (e.g., removing copyrighted or harmful content) and personalization (e.g., adding user-specific characters or styles). Compared to prior datasets used in localization and model editing literature, our probing set is significantly larger in both scale and semantic diversity, enabling a more comprehensive evaluation.

For each target knowledge $\kappa$, we construct a set of knowledge-specific prompts $\{p_1^\kappa, p_2^\kappa, \dots, p_N^\kappa\}$, designed to capture that knowledge in diverse contexts. For example, for $\kappa=$\textit{“the Batman”}, a prompt $p_i^\kappa$ could be \textit{“the Batman walking through a desert”}. To isolate the contribution of the target knowledge $\kappa$ in each prompt $p_i^\kappa$, we also define knowledge-agnostic prompts (denoted as $p_i^{\kappa\text{-neutral}}$) for every knowledge-specific prompt $p_i^\kappa$, where $p_i^{\kappa\text{-neutral}}$ is derived from $p_i^\kappa$ by replacing the target knowledge with a semantically related but generic placeholder (e.g., \textit{“a character walking through a desert”}). These paired prompt sets allow us to perform controlled interventions for evaluating knowledge localization and editing behavior in subsequent sections. For more details on the dataset construction, statistics, and representative prompt examples, please refer to Appendix~\ref{app:probe_dataset}.

\newcommand{\ttoken}{\mathbf{x}}
\newcommand{\itoken}{\mathbf{y}}
\newcommand{\key}{\mathbf{k}}
\newcommand{\query}{\mathbf{q}}
\newcommand{\valuem}{\mathbf{v}}

\subsection{Localization Method}
\label{sec:localization_method}
Our goal is to identify which layers within a DiT-based text-to-image model are responsible for encoding specific semantic knowledge. Specifically, given a prompt $p_i^\kappa$ (e.g., \textit{“Albert Einstein walking in the street”}), where $\kappa$ denotes the target knowledge (i.e., \textit{“Albert Einstein”}), we aim to pinpoint which blocks in the model are primarily responsible for representing $\kappa$. By localizing the internal representation of such knowledge, we can better understand how knowledge is distributed across the model’s architecture and enable targeted interventions such as editing, personalization, or unlearning.


We leverage \textit{attention contribution} \cite{elhage2021mathematical, dar2022analyzing, zarei2024improving} to identify the layers responsible for generating specific content in the image.
At a given layer, the attention contribution of a text token to image tokens quantifies how much that token influences the embeddings of the image tokens.
We localize the layers where a text token exhibits higher attention contribution, interpreting them as the stages where the corresponding style, object, or pattern is synthesized. We adopt attention contribution as our localization signal because it offers an intuitive and principled way to trace how textual information propagates through the model and influences the generated image. Moreover, it can be universally applied across a wide range of DiTs, as it builds on the shared mechanism of attention computation.

More formally, consider layer $\ell$ of a diffusion transformer with $L$ layers equipped with a multi-head cross-attention\footnote{shared attention mechanism in the case of MMDiTs} mechanism comprising $H$ heads. 
For each head $h \in [H]$, let the query, key, value, and output projection matrices be denoted by $W_{q}^h$, $W_{k}^h$, $W_{v}^h$, and $W_{o}^h$, respectively.
Let $\ttoken_1, \ttoken_2, \dots, \ttoken_T$ represent the token embeddings of the input prompt, and $\itoken_1, \itoken_2, \dots, \itoken_I$ denote the token embeddings of the image tokens at layer $\ell$. For head $h$, let the projection of the text token $\ttoken_j$ onto the key and value matrices be denoted by $\key_j^h$ and $\valuem_j^h$, respectively,
and let the projection of the image token $\itoken_i$ onto the query matrix be denoted by $\query_i^h$.
Then, the attention contribution of text token $\ttoken_j$ to image token $\itoken_i$, aggregated over all heads, can be expressed as:
\begin{align*}
\text{cont}_{i, j} = \left\| \sum_{h=1}^H \text{attn}_{i, j}^h\ \valuem_j^h \ W_o^h \right\|_2,
\end{align*}
where $\text{attn}_{i, j}^h$ is the attention weight between image token $\itoken_i$ and text token $\ttoken_j$, computed as:
\begin{align*}
\text{attn}_{i, j}^h = \textsc{Softmax} \left(
\left\{
\frac{
\langle \query_i^h, \key_r^h \rangle
}{\sqrt{d_h}}
\right\}_{r=1}^T
\right)_j,
\end{align*}
where $d_h$ is the head dimensionality, and the softmax is taken over all text tokens for a fixed image token $\itoken_i$.
To compute the overall contribution of a text token $\ttoken_j$, we average over all image tokens.
Finally, for tokens of interest $\{\ttoken_{j_1}, \ttoken_{j_2}, \dots, \ttoken_{j_\tau}\}$ corresponding to the target knowledge $\kappa$ in the prompt $p_i^\kappa$, we compute their attention contribution across all layers $\ell \in [L]$, and identify the layers with the highest aggregated contribution as those most responsible for generating the corresponding style, object, or pattern in the image. 
\begin{figure}[t]
    \centering
    \includegraphics[width=\linewidth]{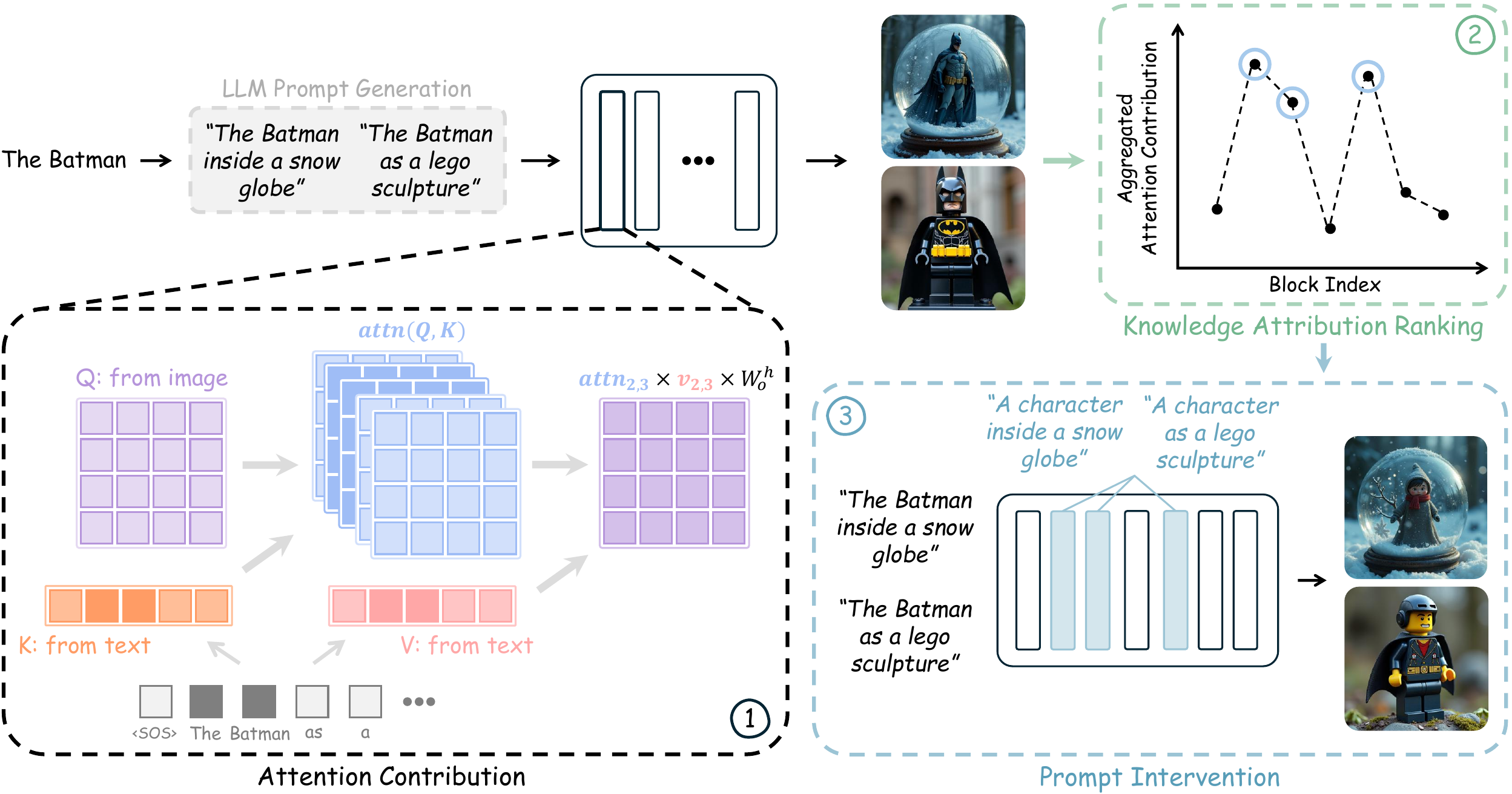}
        \caption{\textbf{Overview of our knowledge localization method.} We first generate images from prompts $\{p_i^\kappa\}$ containing target knowledge $\kappa$, and compute token-level attention contributions across layers. Aggregated scores identify the top-$K$ blocks $\mathcal{B}_K^\kappa$ most responsible for encoding $\kappa$. Replacing their inputs with knowledge-agnostic prompts $\{p_i^{\kappa\text{-neutral}}\}$ suppresses the knowledge in the output.}
    \label{fig:localization_with_attn_cont}
\end{figure}

Figure~\ref{fig:localization_with_attn_cont} illustrates the overall pipeline of our knowledge localization method. Given a target knowledge $\kappa$, we first construct a set of prompts $\{p_1^\kappa, p_2^\kappa, \dots, p_N^\kappa\}$ that contain the knowledge, either manually or using an LLM. Using the DiT model, we generate images and compute the attention contribution of the tokens $\{\ttoken_{j_1}, \ttoken_{j_2}, \dots, \ttoken_{j_\tau}\}$ corresponding to $\kappa$ in each prompt $p_i^\kappa$ at each layer (step 1 in Figure~\ref{fig:localization_with_attn_cont}). These values are averaged across seeds and prompts to obtain a per-layer score indicating how much each block contributes to injecting the knowledge into the image (step 2 in Figure~\ref{fig:localization_with_attn_cont}). We then select the top-$K$ most dominant blocks ($\mathcal{B}_K^\kappa$) as the most informative.

To verify the role of the localized blocks $\mathcal{B}_K^\kappa$, we generate images using the original prompts $\{p_1^\kappa, p_2^\kappa, \dots, p_N^\kappa\}$, but replace the inputs to the $\mathcal{B}_K^\kappa$ with knowledge-agnostic prompts $\{p_1^{\kappa\text{-neutral}}, p_2^{\kappa\text{-neutral}}, \dots, p_N^{\kappa\text{-neutral}}\}$, which omit the knowledge (step 3 in Figure~\ref{fig:localization_with_attn_cont}). In models like PixArt-$\alpha$, this is done by swapping the cross-attention input, and for MMDiT-based models like FLUX, which use a separate prompt branch, we perform two passes, one with $\{p_i^\kappa\}$ and one with $\{p_i^{\kappa\text{-neutral}}\}$, and overwrite the text branch input in the $\mathcal{B}_K^\kappa$ of the first pass with those from the second.

\subsection{Experiments and Results}
\label{sec:method_experiments_and_results}

In this section, we present the results of our proposed knowledge localization method, evaluating its effectiveness across multiple model architectures and diverse knowledge categories.

\textbf{Baselines and Architectures}~~
We evaluate our knowledge localization method on three state-of-the-art models: PixArt-$\alpha$, FLUX, and SANA, covering a range of DiT-based architectural designs. PixArt-$\alpha$  injects prompt information into the image (latent) space via cross-attention blocks using a pretrained T5 encoder \citep{raffel2020exploring}, while SANA uses a lightweight LLM-based encoder \citep{team2024gemma} instead. In contrast, MMDiT-based models such as FLUX maintain a separate prompt branch, parallel to the image branch, which is updated throughout the model and merges with image representations through shared attention layers. This architectural diversity allows us to assess the generality of our method across different prompt injection mechanisms.

\begin{figure}[t]
    \centering
    \includegraphics[width=\linewidth]{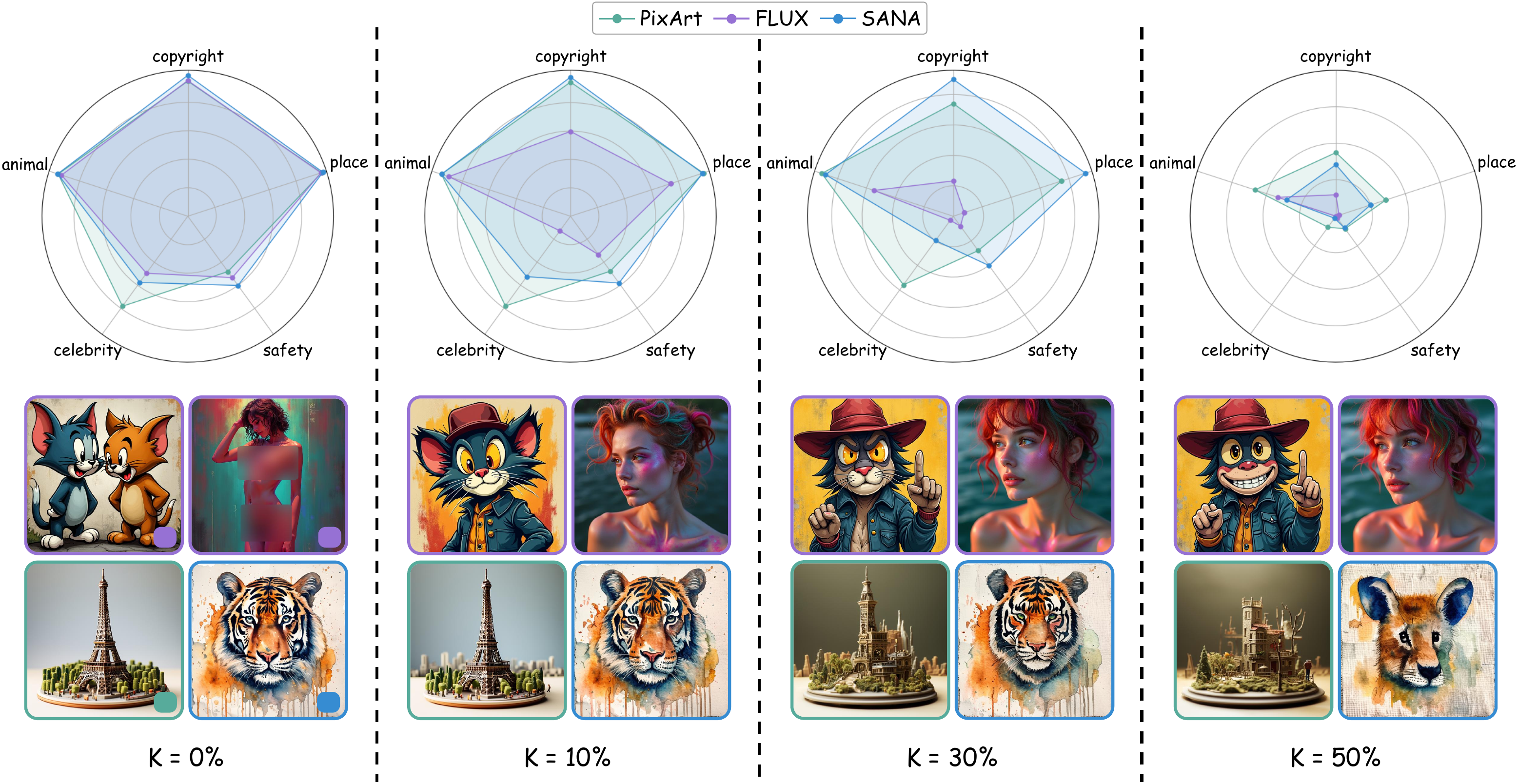}
    \caption{\textbf{Differences in how knowledge is localized across categories and models.} LLaVA-based evaluations and generation samples as the number of intervened blocks $K$ increases, where $K$ denotes the top-$K$ most informative blocks identified by our localization method. Some knowledge types (e.g., copyright) are highly concentrated in a few blocks, while others (e.g., animals) are more distributed across the model. Examples include outputs from the base models and their intervened counterparts.}
    \label{fig:quantitative_and_qualitative_results_of_localization}
\end{figure}

\textbf{Evaluation Metrics}~~ To evaluate the presence of target knowledge in generated images, we use multiple complementary metrics. First, we use CLIP \citep{radford2021learning} to measure the semantic alignment between the target knowledge and the generated image. Second, we leverage LLaVA's \citep{liu2023visual} visual question answering capabilities by explicitly querying whether the knowledge appears in the image. Finally, for style-related concepts, where CLIP and LLaVA are less reliable, we employ the CSD (Contrastive Style Descriptors) \cite{somepalli2024measuring} metric, which is more robust for assessing stylistic consistency.

\textbf{Dataset}~~
We use our proposed dataset, $\datasetname$, spanning all six knowledge categories. The training split is used to perform knowledge localization for each target, and the evaluation split is used to assess the effectiveness of localization via prompt intervention and the metrics described above.

\textbf{Results}~~ 
Figure~\ref{fig:localization_categories_in_different_models} presents the results of our localization method across different model architectures and knowledge categories. For each model, the heatmap bars show how frequently each block is selected among the top-$K$ most informative blocks (with $K=40\%$ of the model’s total blocks), aggregated across knowledge categories. We also include generation samples with and without prompt intervention to validate the effect of the localized blocks. Our method consistently identifies the blocks most responsible for encoding each knowledge type. Notably, we observe that knowledge is distributed quite differently across model architectures. In SANA, knowledge tends to be highly concentrated in a narrow set of blocks, whereas in PixArt-$\alpha$, the distribution is more diffuse—though certain blocks still emerge as consistently dominant. This architectural disparity in how knowledge is stored underscores the importance of localization methods adaptable across architectures, as our approach is—capable of reliably identifying the relevant regions where knowledge is encoded.

Figure~\ref{fig:quantitative_and_qualitative_results_of_localization} illustrates how different knowledge categories are localized across models. In each column, for every target knowledge in our dataset, we first identify the top-$K$ most informative blocks per model using our localization method. We then evaluate the effect of prompt intervention on these blocks using the LLaVA-based evaluation metric, which is shown on a $0-100\%$ scale. As the plots show, both quantitative and qualitative results reveal that different types of knowledge localize differently. Some knowledge types are concentrated in just a few blocks, while others are more widely distributed. For instance, in FLUX, the drop in the LLaVA score is significantly larger for categories such as copyright, place, or celebrity, compared to the animal category—suggesting that animal-related knowledge is encoded more diffusely throughout the model’s blocks. 


\begin{figure}[t]
    \centering
    \includegraphics[width=0.43\linewidth]{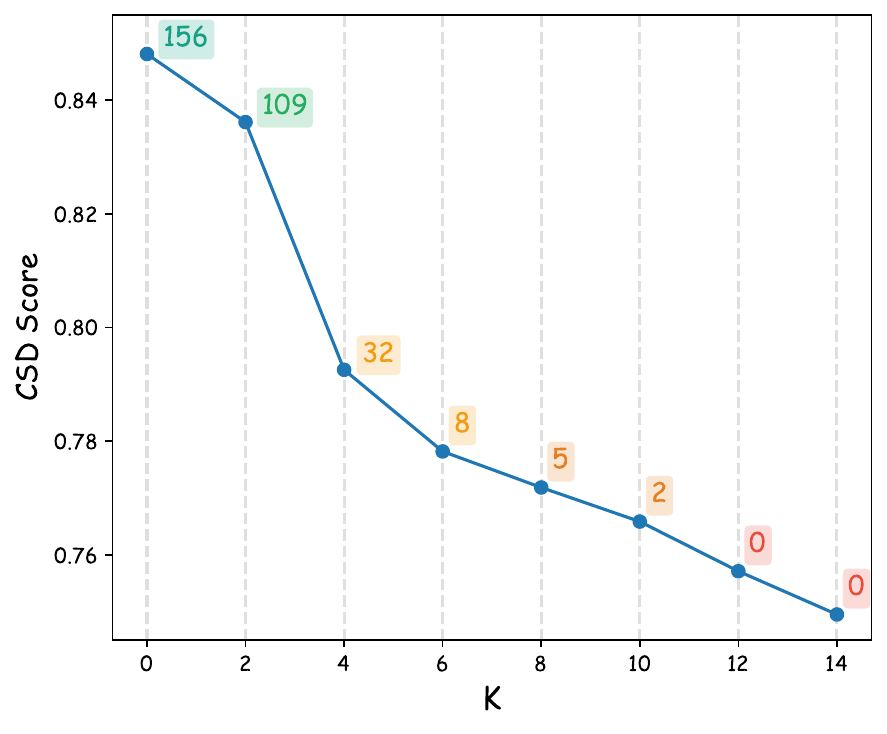}\hfill
    \includegraphics[width=0.56\linewidth]{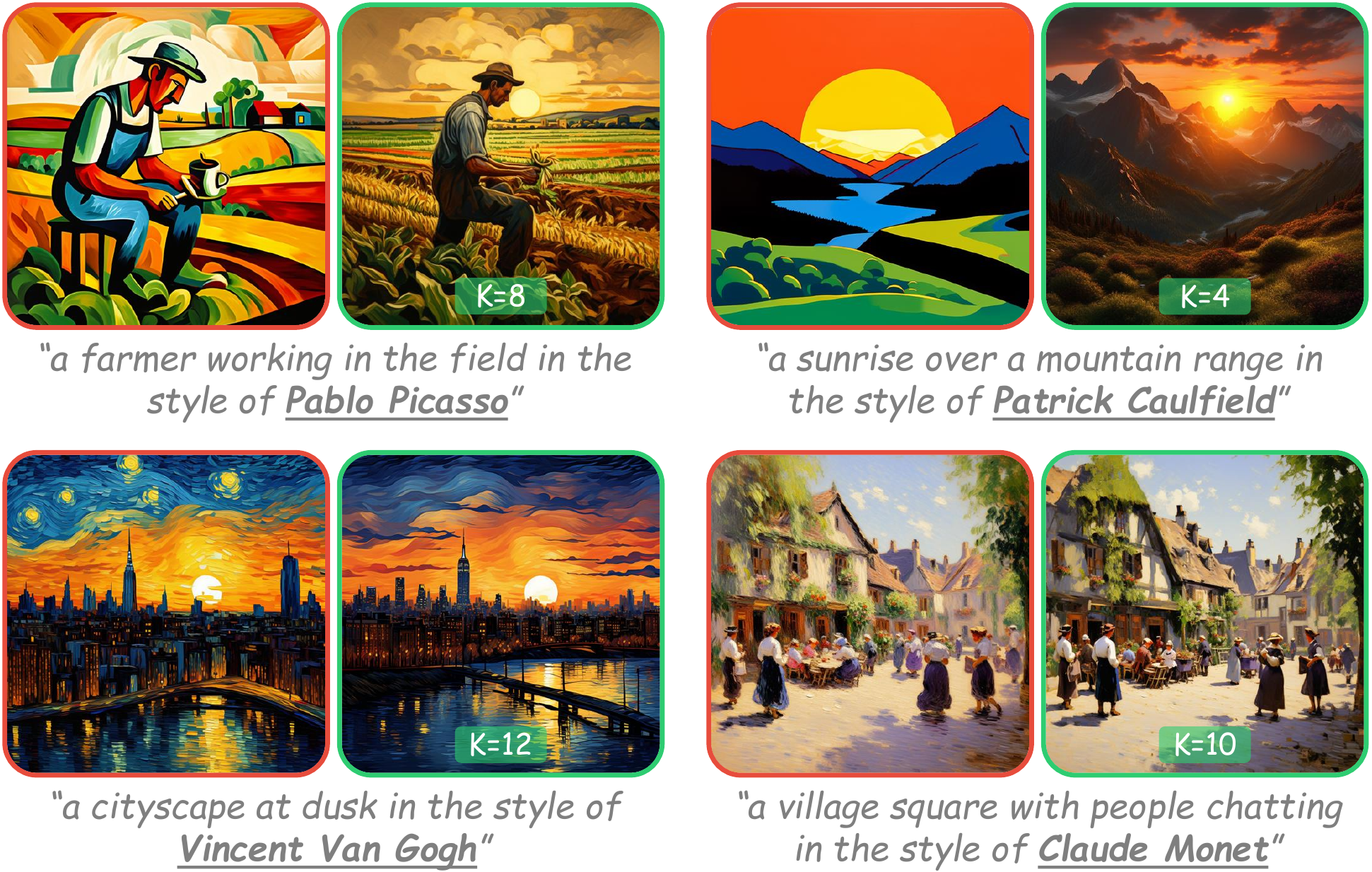}
    \caption{\textbf{Variation in how artistic styles are localized within the model.} We report CSD scores for various artists in the PixArt-$\alpha$ model as the number of intervened blocks $K$ increases. The numbers indicate how many artist styles remain identifiable at each $K$. While styles like \textit{Patrick Caulfield} are localized in fewer blocks, others like \textit{Van Gogh} are distributed more. }
    \label{fig:csd}
\end{figure}

To further examine how knowledge localization varies within a single category, we analyze the differences across individual target knowledge—for example, comparing \textit{“Pablo Picasso”} and \textit{“Van Gogh”} within the artistic style category. Using our localization method, we identify the top-$K$ most informative blocks and evaluate how well the target style is preserved in the generated images under prompt intervention, using the CSD metric and a predefined threshold. Figure~\ref{fig:csd} shows results for the PixArt-$\alpha$ model. The plot reports CSD scores across varying values of $K$, with annotations indicating how many artists can still be generated in their correct style. Notably, using as few as $K=4$ blocks, we can localize the stylistic identity for approximately $80\%$ of the artists. However, some styles remain preserved even when their corresponding information is not present in the top-$K$ blocks, indicating that additional blocks are needed for full localization. On the right side of Figure~\ref{fig:csd}, we show qualitative examples from different artists. As illustrated, styles like \textit{Picasso} are more localized (typically requiring 6–8 blocks), whereas styles like \textit{Van Gogh} are more distributed and require a larger set of blocks (around 12) for effective representation. We further explore whether there is a correlation between the nature of the artistic style (e.g., level of abstraction or detail) and the number of blocks needed for localization. Additional analysis can be found in Appendix~\ref{app:style_correlation_on_localization}.

For more qualitative and quantitative results, see Appendix~\ref{app:localization_qualitative_and_quantitative_results}. Also, to highlight the efficiency and effectiveness of our method, we compare it with a brute-force localization approach in Appendix~\ref{app:comparison_brute_force}.


\section{Applications}
\begin{figure}[t]
    \centering
    \includegraphics[width=\linewidth]{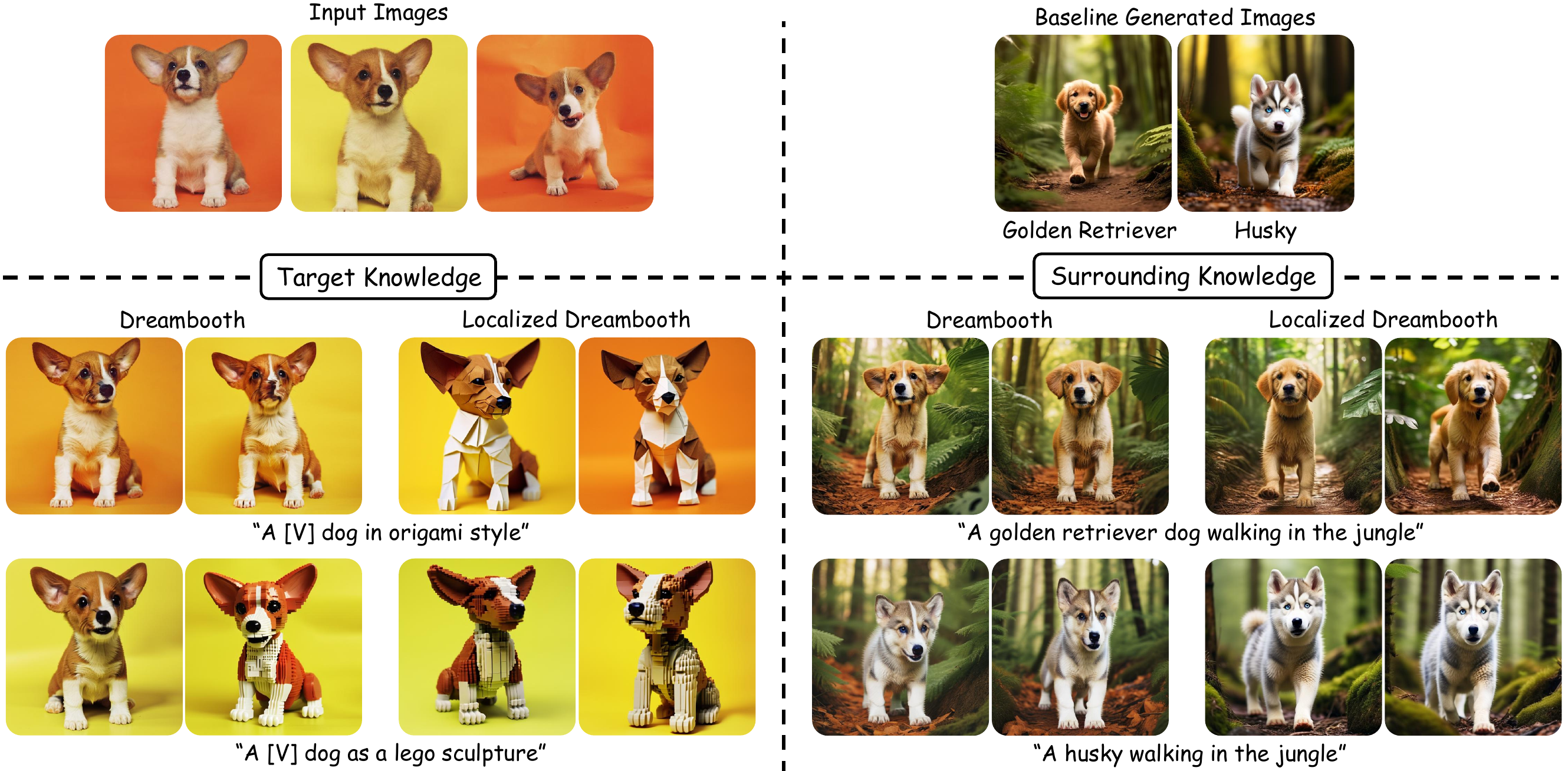}
    \caption{\textbf{Improved prompt alignment and surrounding identity preservation via localized DreamBooth.} Left: Localized fine-tuning better adheres to prompt specifications. Right: Surrounding class-level identities are better preserved, demonstrating reduced interference with other concepts.}
    \label{fig:qualitative_dreambooth}
    \vspace{6pt}
    \includegraphics[width=\linewidth]{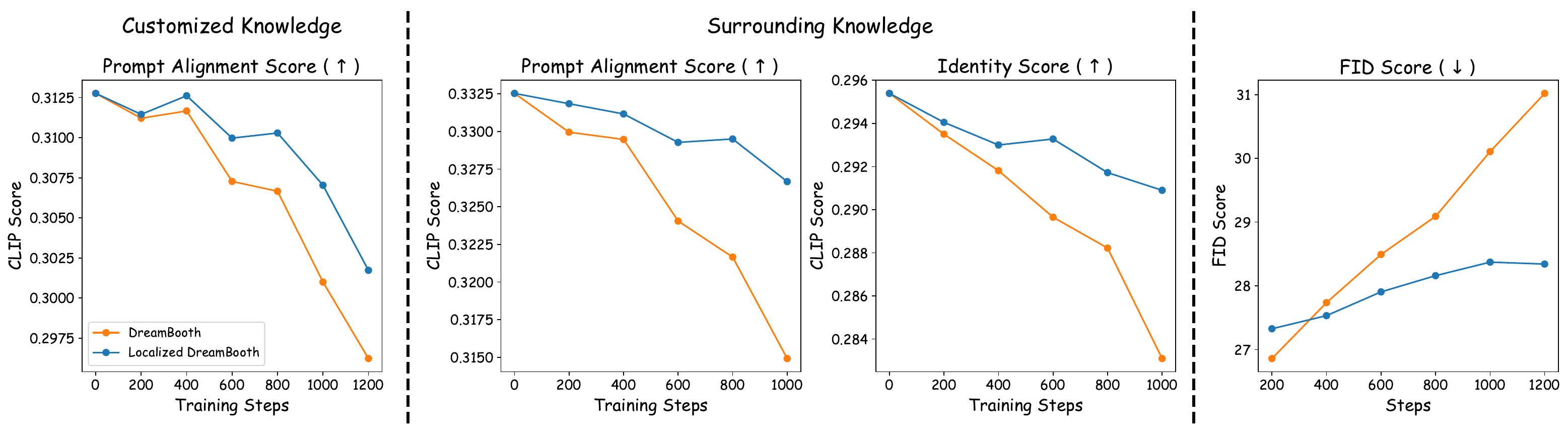}
    \caption{\textbf{Improved quantitative performance with localized DreamBooth.} Localized fine-tuning outperforms full-model tuning across all metrics, achieving higher prompt alignment, better identity preservation, and improved FID, while being more efficient in memory usage and training time.}
    \label{fig:quantitative_dreambooth}
\end{figure}

\subsection{Model Personalization}

Model personalization aims to synthesize high-fidelity images of a subject in novel scenes, poses, colors, and configurations using only a few reference images. We follow the DreamBooth setup \citep{ruiz2023dreambooth}, where a unique identifier token is assigned to the new subject, and the model is fine-tuned for a few epochs to internalize the subject’s visual identity and associate it with that token. For details on the DreamBooth setup and task formulation, please refer to Appendix~\ref{app:model_personalization}.

Unlike conventional DreamBooth, we leverage knowledge localization to precisely guide which parts of the model to fine-tune. Given a new subject, we first infer its semantic class (e.g., dog for a specific dog instance), then use our method to identify the blocks most responsible for encoding knowledge related to that class. Fine-tuning is then restricted to only those blocks. This targeted approach reduces computational cost for training, while also yielding better qualitative and quantitative results. Our method leads to stronger preservation of surrounding concepts and scene consistency, and exhibits superior prompt alignment in novel scenarios compared to full-model fine-tuning (see Section \ref{sec:application_experiments_and_results}). 

\subsection{Concept Unlearning}

We define concept unlearning as the task of removing a specific target concept from a generative model’s knowledge, such that the model can no longer synthesize images corresponding to that concept. Rather than retraining the model from scratch on a dataset with the concept manually excluded, our goal is to achieve this effect through minimal and targeted intervention. To this end, we follow the setup proposed by \citet{kumari2023ablating}. For a given target concept (e.g., “The Batman”), we assume access to an associated anchor concept (e.g., “a character”)—a broader or semantically related category that serves as a neutral substitute. The objective is to align the model’s output distribution for the target concept with that of the anchor, effectively erasing the specific knowledge while preserving the model’s general generative capabilities. For more details on the setup, see Appendix~\ref{app:concept_unlearning}.

As with model personalization, we incorporate \textit{concept localization} to identify which blocks in the model encode information related to the target concept. Rather than updating the entire model, we restrict fine-tuning to these localized regions. This targeted unlearning not only improves memory efficiency and speeds up training, but also leads to comparable or improved results in both qualitative and quantitative evaluations, as we will demonstrate in the following section.

\begin{figure}[t]
    \centering
    \includegraphics[width=\linewidth]{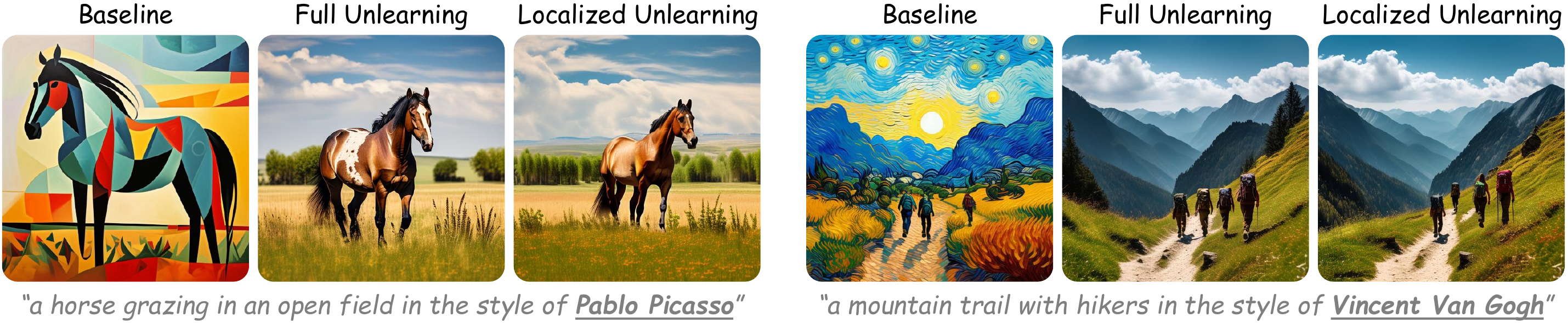}
    \caption{\textbf{Effective and efficient concept unlearning with localized fine-tuning.} Our method removes targeted styles as effectively as full-model tuning, while requiring much less computation.}
    \label{fig:quantitative_unlearning}
    \vspace{6pt}
    
    \includegraphics[width=\linewidth]{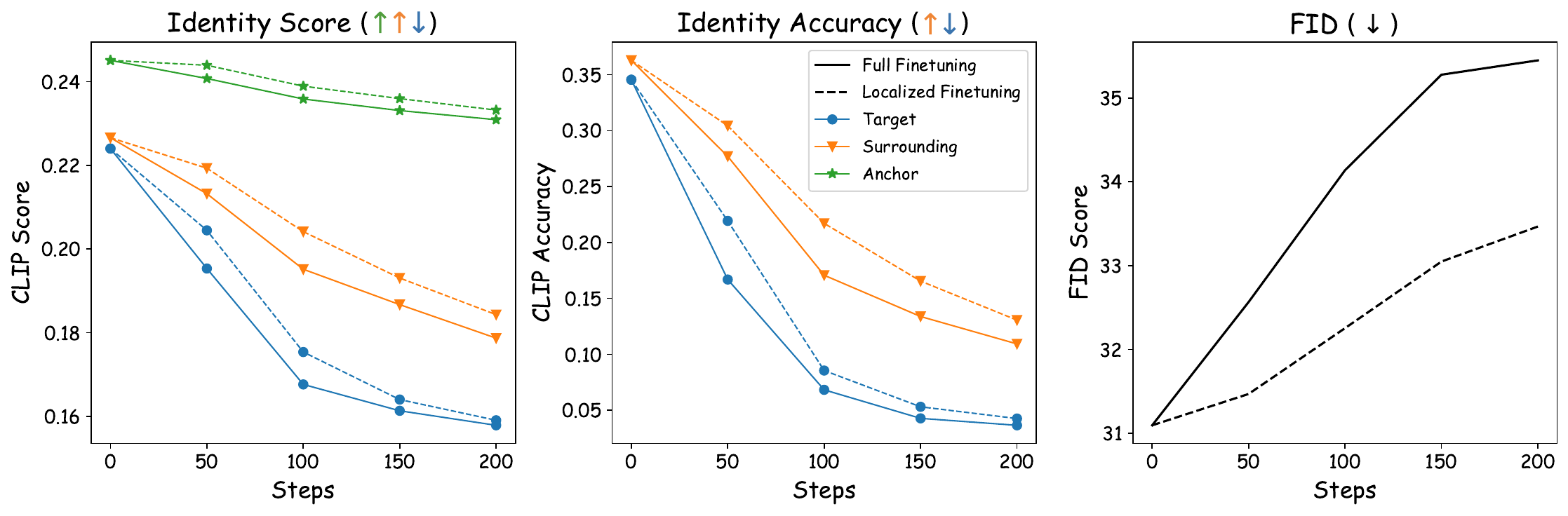}
    \caption{\textbf{Quantitative results showing better performance with localized unlearning.} Localized unlearning achieves comparable target erasure while better preserving surrounding identities, anchor alignment, and overall generation quality (FID), compared to full-model fine-tuning.}
    \label{fig:quantitative_unlearning}
    \vspace{-2pt}
    
\end{figure}

\subsection{Experiments and Results}
\label{sec:application_experiments_and_results}
In this section, we present the results of our proposed localized personalization and unlearning methods, evaluating their effectiveness across both qualitative and quantitative metrics.

\textbf{Setup}~~
We base our experiments on the publicly available PixArt-$\alpha$ \citep{chen2023pixart} model. For model personalization, we follow the setup introduced in DreamBooth \citep{ruiz2023dreambooth}, and apply localized fine-tuning by updating only $K=9$ out of the model’s $28$ transformer blocks. For concept unlearning, we adopt the setup proposed by \citet{kumari2023ablating}, and apply our method by fine-tuning only $K=5$ blocks. We focus primarily on the style category, selecting the 30 artists the model is best at producing—based on CSD scores—and apply unlearning to each. Further experimental details are provided in Appendix~\ref{app:application_experiment}.

\textbf{Evaluation Metrics}~~
For personalized models, we employ multiple metrics to comprehensively assess their performance. \textit{Prompt Alignment Score} is measured via the CLIP Score between the generated image and the corresponding prompt, capturing how well the image reflects the intended scene, style, and semantics described by the prompt. \textit{Identity Score}, also based on CLIP similarity, measures how well the generated image preserves the subject’s visual identity, e.g., retaining the distinctive appearance of a Husky or Golden Retriever when fine-tuning on a specific dog. These surrounding class-level concepts are selected from LLM-generated prompts that resemble the subject's broader category.  Finally, to assess overall image quality, we compute the Fréchet Inception Distance (FID) \cite{heusel2017gans} on a 10k sample subset of the COCO dataset \cite{lin2014microsoft} throughout the fine-tuning process. 

For concept unlearning, we follow the evaluation protocol of \citet{kumari2023ablating}. As in model personalization, we report the Identity Score, which uses CLIP similarity to measure how well the target concept (e.g., an artist’s style) is removed from the generated image. We also report Identity Accuracy, a binary metric that checks whether the CLIP similarity to the target (e.g., “Van Gogh style”) falls below that of the anchor (e.g., “a painting”). Additionally, we compute FID to evaluate the preservation of overall generation quality.

\textbf{Results}~~
As for model personalization, Figure~\ref{fig:quantitative_dreambooth} shows that our targeted fine-tuning consistently outperforms full fine-tuning across all metrics. In terms of prompt adherence, qualitative results (Figure~\ref{fig:qualitative_dreambooth}, left side) show that our method more faithfully reflects user prompts such as \textit{“a [V] dog in origami style”}. On the right, we observe that the identities of surrounding concepts (e.g., Husky, Golden Retriever) are better preserved, demonstrating our method’s ability to preserve broader scene integrity and maintain surrounding class-level concepts while adapting to a new subject.

As for concept unlearning, Figure~\ref{fig:quantitative_unlearning} shows that our localized unlearning approach better preserves the identity of surrounding concepts and maintains alignment with the anchor prompts, while achieving comparable erasure performance on the target concept (see results at 200 steps). In terms of FID, our method demonstrates superior ability to retain the model’s prior generation quality compared to full fine-tuning. Moreover, as illustrated in Figure~\ref{fig:quantitative_unlearning}, our method effectively removes the targeted styles with performance on par with full-model fine-tuning—yet with significantly lower computational cost (15–20\% speedup and approximately 30\% reduction in memory usage).

\section{Conclusion}
In this paper, we presented a model- and knowledge-agnostic method for localizing where specific knowledge resides within the blocks of Diffusion Transformers. Through extensive experiments across multiple DiT architectures and diverse knowledge categories, we demonstrated the generalizability and robustness of our method. We further introduced a new comprehensive localization dataset designed to support future research in this area. Building on our localization, we applied our method to practical downstream tasks, showing that localized fine-tuning improves task-specific performance while being less disruptive to unrelated model behavior and being more efficient. We hope this work serves as a foundation for more interpretable, controllable, and efficient adaptation of DiTs.

\section*{Acknowledgement}
This project was supported in part by a grant from an NSF CAREER AWARD 1942230, ONR YIP award N00014-22-1-2271, ARO’s Early Career Program Award 310902-00001, Army Grant No. W911NF2120076, the NSF award CCF2212458, NSF Award No. 2229885 (NSF Institute for Trustworthy AI in Law and Society, TRAILS), a MURI grant 14262683, an award from meta 314593-00001 and an award from Capital One.

\bibliographystyle{abbrvnat}
\bibliography{ref}

@inproceedings{ronneberger2015u,
  title={U-net: Convolutional networks for biomedical image segmentation},
  author={Ronneberger, Olaf and Fischer, Philipp and Brox, Thomas},
  booktitle={Medical image computing and computer-assisted intervention--MICCAI 2015: 18th international conference, Munich, Germany, October 5-9, 2015, proceedings, part III 18},
  pages={234--241},
  year={2015},
  organization={Springer}
}

@inproceedings{peebles2023scalable,
  title={Scalable diffusion models with transformers},
  author={Peebles, William and Xie, Saining},
  booktitle={Proceedings of the IEEE/CVF international conference on computer vision},
  pages={4195--4205},
  year={2023}
}

@article{zarei2024improving,
  title={Improving Compositional Attribute Binding in Text-to-Image Generative Models via Enhanced Text Embeddings},
  author={Zarei, Arman and Rezaei, Keivan and Basu, Samyadeep and Saberi, Mehrdad and Moayeri, Mazda and Kattakinda, Priyatham and Feizi, Soheil},
  journal={arXiv preprint arXiv:2406.07844},
  year={2024}
}

@inproceedings{
basu2024localizing,
title={Localizing and Editing Knowledge In Text-to-Image Generative Models},
author={Samyadeep Basu and Nanxuan Zhao and Vlad I Morariu and Soheil Feizi and Varun Manjunatha},
booktitle={The Twelfth International Conference on Learning Representations},
year={2024},
url={https://openreview.net/forum?id=Qmw9ne6SOQ}
}

@misc{avrahami2025stableflowvitallayers,
      title={Stable Flow: Vital Layers for Training-Free Image Editing}, 
      author={Omri Avrahami and Or Patashnik and Ohad Fried and Egor Nemchinov and Kfir Aberman and Dani Lischinski and Daniel Cohen-Or},
      year={2025},
      eprint={2411.14430},
      archivePrefix={arXiv},
      primaryClass={cs.CV},
      url={https://arxiv.org/abs/2411.14430}, 
}

@misc{helbling2025conceptattentiondiffusiontransformerslearn,
      title={ConceptAttention: Diffusion Transformers Learn Highly Interpretable Features}, 
      author={Alec Helbling and Tuna Han Salih Meral and Ben Hoover and Pinar Yanardag and Duen Horng Chau},
      year={2025},
      eprint={2502.04320},
      archivePrefix={arXiv},
      primaryClass={cs.CV},
      url={https://arxiv.org/abs/2502.04320}, 
}

@inproceedings{
basu2024on,
title={On Mechanistic Knowledge Localization in Text-to-Image Generative Models},
author={Samyadeep Basu and Keivan Rezaei and Priyatham Kattakinda and Vlad I Morariu and Nanxuan Zhao and Ryan A. Rossi and Varun Manjunatha and Soheil Feizi},
booktitle={Forty-first International Conference on Machine Learning},
year={2024},
url={https://openreview.net/forum?id=fsVBsxjRER}
}

@article{elhage2021mathematical,
   title={A Mathematical Framework for Transformer Circuits},
   author={Elhage, Nelson and Nanda, Neel and Olsson, Catherine and Henighan, Tom and Joseph, Nicholas and Mann, Ben and Askell, Amanda and Bai, Yuntao and Chen, Anna and Conerly, Tom and DasSarma, Nova and Drain, Dawn and Ganguli, Deep and Hatfield-Dodds, Zac and Hernandez, Danny and Jones, Andy and Kernion, Jackson and Lovitt, Liane and Ndousse, Kamal and Amodei, Dario and Brown, Tom and Clark, Jack and Kaplan, Jared and McCandlish, Sam and Olah, Chris},
   year={2021},
   journal={Transformer Circuits Thread},
   note={https://transformer-circuits.pub/2021/framework/index.html}
}

@article{dar2022analyzing,
  title={Analyzing transformers in embedding space},
  author={Dar, Guy and Geva, Mor and Gupta, Ankit and Berant, Jonathan},
  journal={arXiv preprint arXiv:2209.02535},
  year={2022}
}

@article{vaswani2017attention,
  title={Attention is all you need},
  author={Vaswani, Ashish and Shazeer, Noam and Parmar, Niki and Uszkoreit, Jakob and Jones, Llion and Gomez, Aidan N and Kaiser, {\L}ukasz and Polosukhin, Illia},
  journal={Advances in neural information processing systems},
  volume={30},
  year={2017}
}

@article{ho2020denoising,
  title={Denoising diffusion probabilistic models},
  author={Ho, Jonathan and Jain, Ajay and Abbeel, Pieter},
  journal={Advances in neural information processing systems},
  volume={33},
  pages={6840--6851},
  year={2020}
}

@inproceedings{nichol2021improved,
  title={Improved denoising diffusion probabilistic models},
  author={Nichol, Alexander Quinn and Dhariwal, Prafulla},
  booktitle={International conference on machine learning},
  pages={8162--8171},
  year={2021},
  organization={PMLR}
}

@inproceedings{rombach2022high,
  title={High-resolution image synthesis with latent diffusion models},
  author={Rombach, Robin and Blattmann, Andreas and Lorenz, Dominik and Esser, Patrick and Ommer, Bj{\"o}rn},
  booktitle={Proceedings of the IEEE/CVF conference on computer vision and pattern recognition},
  pages={10684--10695},
  year={2022}
}

@article{ho2022classifier,
  title={Classifier-free diffusion guidance},
  author={Ho, Jonathan and Salimans, Tim},
  journal={arXiv preprint arXiv:2207.12598},
  year={2022}
}

@misc{flux2024,
    author={Black Forest Labs},
    title={FLUX},
    year={2024},
    howpublished={\url{https://github.com/black-forest-labs/flux}},
}

@inproceedings{esser2024scaling,
  title={Scaling rectified flow transformers for high-resolution image synthesis},
  author={Esser, Patrick and Kulal, Sumith and Blattmann, Andreas and Entezari, Rahim and M{\"u}ller, Jonas and Saini, Harry and Levi, Yam and Lorenz, Dominik and Sauer, Axel and Boesel, Frederic and others},
  booktitle={Forty-first international conference on machine learning},
  year={2024}
}

@article{lipman2022flow,
  title={Flow matching for generative modeling},
  author={Lipman, Yaron and Chen, Ricky TQ and Ben-Hamu, Heli and Nickel, Maximilian and Le, Matt},
  journal={arXiv preprint arXiv:2210.02747},
  year={2022}
}

@article{hertz2022prompt,
  title={Prompt-to-prompt image editing with cross attention control},
  author={Hertz, Amir and Mokady, Ron and Tenenbaum, Jay and Aberman, Kfir and Pritch, Yael and Cohen-Or, Daniel},
  journal={arXiv preprint arXiv:2208.01626},
  year={2022}
}

@inproceedings{tumanyan2023plug,
  title={Plug-and-play diffusion features for text-driven image-to-image translation},
  author={Tumanyan, Narek and Geyer, Michal and Bagon, Shai and Dekel, Tali},
  booktitle={Proceedings of the IEEE/CVF Conference on Computer Vision and Pattern Recognition},
  pages={1921--1930},
  year={2023}
}

@inproceedings{kumari2023ablating,
  title={Ablating concepts in text-to-image diffusion models},
  author={Kumari, Nupur and Zhang, Bingliang and Wang, Sheng-Yu and Shechtman, Eli and Zhang, Richard and Zhu, Jun-Yan},
  booktitle={Proceedings of the IEEE/CVF International Conference on Computer Vision},
  pages={22691--22702},
  year={2023}
}

@article{zarei2024understanding,
  title={Understanding and Mitigating Compositional Issues in Text-to-Image Generative Models},
  author={Zarei, Arman and Rezaei, Keivan and Basu, Samyadeep and Saberi, Mehrdad and Moayeri, Mazda and Kattakinda, Priyatham and Feizi, Soheil},
  journal={arXiv preprint arXiv:2406.07844},
  year={2024}
}

@article{xie2024sana,
  title={Sana: Efficient high-resolution image synthesis with linear diffusion transformers},
  author={Xie, Enze and Chen, Junsong and Chen, Junyu and Cai, Han and Tang, Haotian and Lin, Yujun and Zhang, Zhekai and Li, Muyang and Zhu, Ligeng and Lu, Yao and others},
  journal={arXiv preprint arXiv:2410.10629},
  year={2024}
}

@article{chen2023pixart,
  title={Pixart-$\alpha$: Fast training of diffusion transformer for photorealistic text-to-image synthesis},
  author={Chen, Junsong and Yu, Jincheng and Ge, Chongjian and Yao, Lewei and Xie, Enze and Wu, Yue and Wang, Zhongdao and Kwok, James and Luo, Ping and Lu, Huchuan and others},
  journal={arXiv preprint arXiv:2310.00426},
  year={2023}
}

@inproceedings{ruiz2023dreambooth,
  title={Dreambooth: Fine tuning text-to-image diffusion models for subject-driven generation},
  author={Ruiz, Nataniel and Li, Yuanzhen and Jampani, Varun and Pritch, Yael and Rubinstein, Michael and Aberman, Kfir},
  booktitle={Proceedings of the IEEE/CVF conference on computer vision and pattern recognition},
  pages={22500--22510},
  year={2023}
}

@inproceedings{lin2014microsoft,
  title={Microsoft coco: Common objects in context},
  author={Lin, Tsung-Yi and Maire, Michael and Belongie, Serge and Hays, James and Perona, Pietro and Ramanan, Deva and Doll{\'a}r, Piotr and Zitnick, C Lawrence},
  booktitle={Computer vision--ECCV 2014: 13th European conference, zurich, Switzerland, September 6-12, 2014, proceedings, part v 13},
  pages={740--755},
  year={2014},
  organization={Springer}
}

@article{heusel2017gans,
  title={Gans trained by a two time-scale update rule converge to a local nash equilibrium},
  author={Heusel, Martin and Ramsauer, Hubert and Unterthiner, Thomas and Nessler, Bernhard and Hochreiter, Sepp},
  journal={Advances in neural information processing systems},
  volume={30},
  year={2017}
}

@article{somepalli2024measuring,
  title={Measuring style similarity in diffusion models},
  author={Somepalli, Gowthami and Gupta, Anubhav and Gupta, Kamal and Palta, Shramay and Goldblum, Micah and Geiping, Jonas and Shrivastava, Abhinav and Goldstein, Tom},
  journal={arXiv preprint arXiv:2404.01292},
  year={2024}
}

@inproceedings{cao2023masactrl,
  title={Masactrl: Tuning-free mutual self-attention control for consistent image synthesis and editing},
  author={Cao, Mingdeng and Wang, Xintao and Qi, Zhongang and Shan, Ying and Qie, Xiaohu and Zheng, Yinqiang},
  booktitle={Proceedings of the IEEE/CVF international conference on computer vision},
  pages={22560--22570},
  year={2023}
}

@inproceedings{liu2024towards,
  title={Towards understanding cross and self-attention in stable diffusion for text-guided image editing},
  author={Liu, Bingyan and Wang, Chengyu and Cao, Tingfeng and Jia, Kui and Huang, Jun},
  booktitle={Proceedings of the IEEE/CVF conference on computer vision and pattern recognition},
  pages={7817--7826},
  year={2024}
}

@article{yoon2024safree,
  title={Safree: Training-free and adaptive guard for safe text-to-image and video generation},
  author={Yoon, Jaehong and Yu, Shoubin and Patil, Vaidehi and Yao, Huaxiu and Bansal, Mohit},
  journal={arXiv preprint arXiv:2410.12761},
  year={2024}
}

@inproceedings{gandikota2024unified,
  title={Unified concept editing in diffusion models},
  author={Gandikota, Rohit and Orgad, Hadas and Belinkov, Yonatan and Materzy{\'n}ska, Joanna and Bau, David},
  booktitle={Proceedings of the IEEE/CVF Winter Conference on Applications of Computer Vision},
  pages={5111--5120},
  year={2024}
}

@inproceedings{schramowski2023safe,
  title={Safe latent diffusion: Mitigating inappropriate degeneration in diffusion models},
  author={Schramowski, Patrick and Brack, Manuel and Deiseroth, Bj{\"o}rn and Kersting, Kristian},
  booktitle={Proceedings of the IEEE/CVF Conference on Computer Vision and Pattern Recognition},
  pages={22522--22531},
  year={2023}
}

@article{raffel2020exploring,
  title={Exploring the limits of transfer learning with a unified text-to-text transformer},
  author={Raffel, Colin and Shazeer, Noam and Roberts, Adam and Lee, Katherine and Narang, Sharan and Matena, Michael and Zhou, Yanqi and Li, Wei and Liu, Peter J},
  journal={Journal of machine learning research},
  volume={21},
  number={140},
  pages={1--67},
  year={2020}
}

@article{team2024gemma,
  title={Gemma 2: Improving open language models at a practical size},
  author={Team, Gemma and Riviere, Morgane and Pathak, Shreya and Sessa, Pier Giuseppe and Hardin, Cassidy and Bhupatiraju, Surya and Hussenot, L{\'e}onard and Mesnard, Thomas and Shahriari, Bobak and Ram{\'e}, Alexandre and others},
  journal={arXiv preprint arXiv:2408.00118},
  year={2024}
}

@article{liu2023visual,
  title={Visual instruction tuning},
  author={Liu, Haotian and Li, Chunyuan and Wu, Qingyang and Lee, Yong Jae},
  journal={Advances in neural information processing systems},
  volume={36},
  pages={34892--34916},
  year={2023}
}

@inproceedings{radford2021learning,
  title={Learning transferable visual models from natural language supervision},
  author={Radford, Alec and Kim, Jong Wook and Hallacy, Chris and Ramesh, Aditya and Goh, Gabriel and Agarwal, Sandhini and Sastry, Girish and Askell, Amanda and Mishkin, Pamela and Clark, Jack and others},
  booktitle={International conference on machine learning},
  pages={8748--8763},
  year={2021},
  organization={PmLR}
}

@article{hurst2024gpt,
  title={Gpt-4o system card},
  author={Hurst, Aaron and Lerer, Adam and Goucher, Adam P and Perelman, Adam and Ramesh, Aditya and Clark, Aidan and Ostrow, AJ and Welihinda, Akila and Hayes, Alan and Radford, Alec and others},
  journal={arXiv preprint arXiv:2410.21276},
  year={2024}
}

@misc{wikiart,
  author       = {{WikiArt}},
  title        = {{WikiArt: Visual Art Encyclopedia}},
  year         = {n.d.},
  url         = {https://www.wikiart.org/},
}

\newpage

\appendix

\section{Related Works}
\subsection{Diffusion and Flow Models}
\label{app:related_works_diffusion_and_flow}

Diffusion models belong to a class of generative models based on stochastic differential equations (SDE). The central idea is to progressively add noise to the original data through a stochastic forward process, eventually transforming the data distribution into a simple Gaussian distribution. This forward process is mathematically expressed as 
$$dx = f(x, t) dt + g(t)dW_t,$$ 
where $f(x, t)$ denotes the drift term, $g(t)$ is the diffusion coefficient, and $dW_t$ represents the Wiener process (the infinitesimal increment of standard Brownian motion at time $t$, intuitively understood as an instantaneous Gaussian random perturbation). The reverse process, which aims at reconstructing the original data distribution from noise, is formulated as 
$$dx = [f(x, t) - g^2(t)\nabla_x\log p_t(x)]dt + g(t)dW_t.$$ 
Here, the term $\nabla_x \log p_t(x)$, called the score function, describes the gradient of the data distribution at time $t$. The model is trained to approximate this score function by minimizing the score matching loss, formulated as:
$$\mathbb{E}_{t \sim U(0, T), x \sim p_t(x)}[\lambda(t)||\nabla_x\log p_t(x)-s_\theta(x,t)||^2],$$
where $s_\theta(x, t)$ is a parameterized neural network and $\lambda(t)$ is a time-dependent weighting function. 

Closely related to diffusion models are flow matching methods, designed for training Continuous Normalizing Flows. Flow matching aims to deterministically map an initial noise distribution to a target data distribution via an ordinary differential equation (ODE). The trajectory is determined by a learned vector field described as: 
$$\frac{dx}{dt}=v_\theta(x,t),$$
where $v_\theta(x,t)$ is the parameterized vector field to be trained. The training objective involves minimizing the loss function:
$$\mathbb{E}_{t \sim U(0, T), x \sim p_t(x)}[|v_\theta(x,t)-v_t(x)|^2],$$
where $v_t(x)$ represents the target vector field, and $p_t(x)$ denotes intermediate distributions along the path from the initial to the final data distribution. Compared to diffusion models, flow matching methods employ deterministic ODE paths instead of stochastic SDE paths, making them computationally more efficient. Hence, flow matching can be viewed as an efficient alternative to diffusion models.

\section{Localizing Knowledge in Diffusion Transformers}

\subsection{Probe Dataset Description}
\label{app:probe_dataset}

In this section, we describe the construction of our proposed dataset, $\datasetname$ (\underline{$\mathcal{L}$oc}alization of \underline{$\mathcal{K}$}nowledge), which is organized around six distinct categories of knowledge and concepts: artistic styles (e.g., \textit{“style of Van Gogh”}), celebrities (e.g., \textit{“Albert Einstein”}), sensitive or safety-related content (e.g., \textit{“a naked woman”, “a dead body covered in blood”}), copyrighted characters (e.g., \textit{“the Batman”}), famous landmarks (e.g., \textit{“the Eiffel Tower”}), and animals (e.g., \textit{“a black panther”}). These categories were selected to span a diverse range of visual and semantic information while reflecting key use cases in model unlearning (e.g., removing copyrighted or harmful content) and personalization (e.g., adding user-specific characters or styles).

To construct the target knowledge samples: for the artistic style category, we selected 1,108 samples from the WikiArt Artists dataset \cite{wikiart}. For the remaining categories, we used ChatGPT-4o \cite{hurst2024gpt} to generate a list of representative examples, initialized through a few-shot prompting setup. Prompt augmentation was similarly performed using ChatGPT-4o. For each category, we provided several examples and asked the model to generate diverse, semantically meaningful prompts corresponding to target knowledge instances. Table~\ref{tab:dataset_stats} provides statistics for each category, including the number of target knowledge entries, number of augmentation prompts, and total dataset size. Table~\ref{tab:dataset_prompt_examples} also presents examples of prompts across the six categories. Compared to prior datasets used in localization and model editing, $\datasetname$ is substantially larger in both scale and semantic diversity, facilitating a more comprehensive and rigorous evaluation of knowledge localization methods.

\begin{table}[t]
\centering
\caption{Dataset statistics across six knowledge categories in $\datasetname$}
\begin{tabular}{lcccccc}
\toprule
 & \textbf{Style} & \textbf{Copyright} & \textbf{Safety} & \textbf{Celebrity} & \textbf{Place} & \textbf{Animal} \\
\midrule
\# Target Knowledge     & 1108 & 31 & 50 & 30 & 20 & 40 \\
\# Train Prompts        & 20 & 20 & 10 & 20 & 10 & 20 \\
\# Eval Prompts         & 30 & 30 & 20 & 25 & 20 & 30 \\ \hdashline
&&&&&&\\[-8pt]
Dataset Size & 55400 & 1550 & 1500 & 1350 & 600 & 2000 \\ 
\bottomrule
\end{tabular}
\label{tab:dataset_stats}
\end{table}



\subsection{Comparison with Brute-Force Localization}
\label{app:comparison_brute_force}

To demonstrate the robustness, efficiency, and effectiveness of our localization method, we compared it against a brute-force baseline. Specifically, we implemented a brute-force approach that exhaustively evaluates all possible contiguous block windows of size $K$ within the model. For each candidate window, we applied prompt intervention and evaluated the results using the CLIP score and the CSD score, as described in Section~\ref{sec:method_experiments_and_results}. In the case of the PixArt model, where the number of blocks is 28, the brute-force method explores 28 possible windows (including circular windows), making it computationally expensive.

We set $K=6$ and computed both CLIP and CSD scores for each window. For comparison, we also ran our proposed localization method under the same settings ($K=6$). The brute-force method, when selecting the optimal window for removing style information, resulted in a CLIP score drop of 0.0232 (from 0.2255 to 0.2023) and a CSD score drop of 0.0812 (from 0.8481 to 0.7669). In contrast, our method achieved a CLIP score drop of only 0.0194 (from 0.2255 to 0.2061) and a CSD score drop of 0.0700 (from 0.8481 to 0.7781).

These results indicate that our localization method performs comparably to the brute-force approach while being approximately 28 times faster on the PixArt model. More generally, for a DiT model with $B$ blocks, our method offers a $B\times$ speedup. This highlights both the efficiency and reliability of our approach.

\begin{figure}[t]
    \centering
    \includegraphics[width=\linewidth]{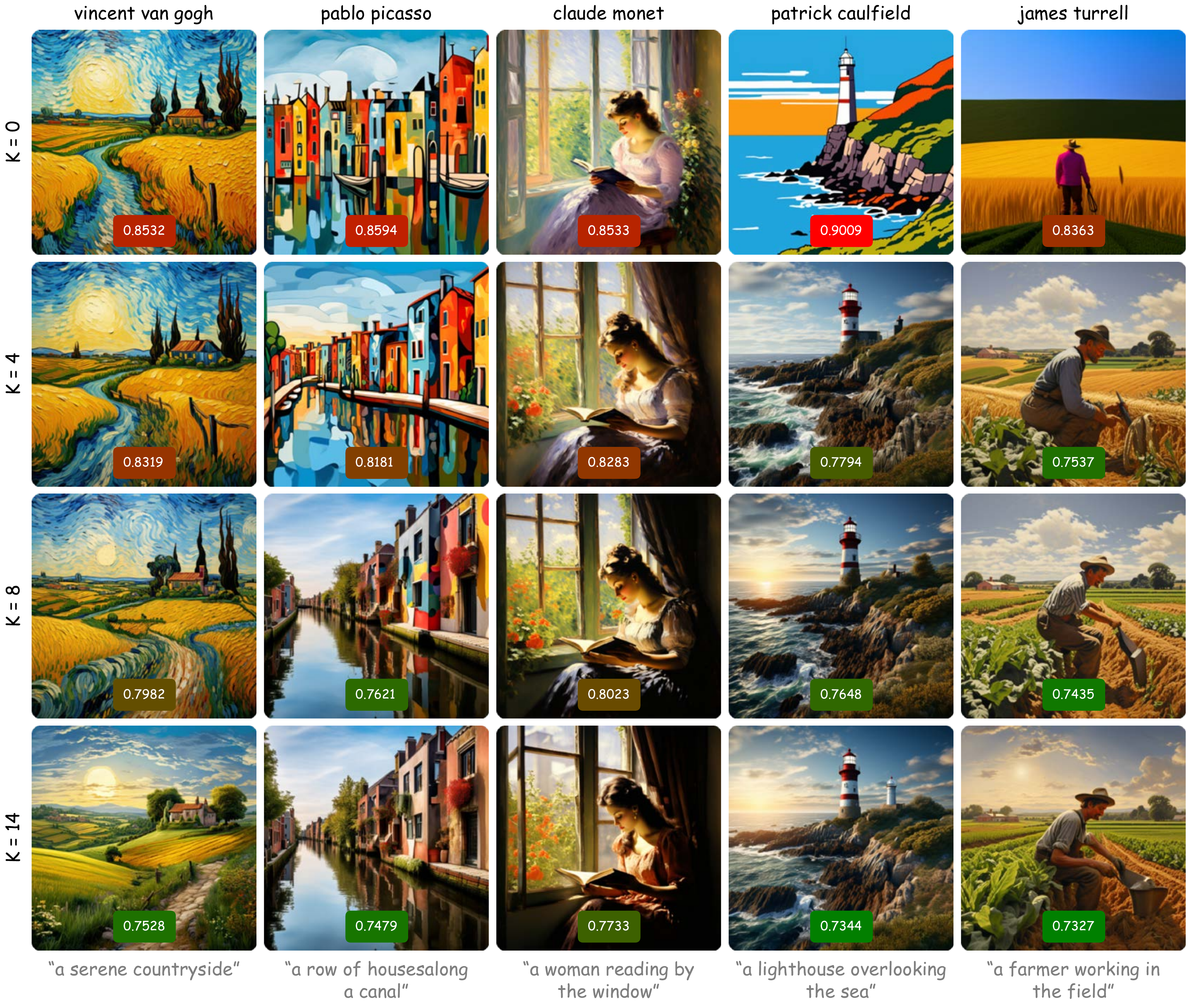}
    \caption{\textbf{Qualitative results of artistic style localization across different values of $K$.} Each column shows generations from the PixArt-$\alpha$ model for a specific artist, comparing outputs with and without prompt intervention on the top-$K$ localized blocks. The colored labels indicate the average CSD distance using a red-to-green spectrum: more reddish colors signify higher similarity to the original style (i.e., style is still preserved), while more greenish colors indicate greater deviation (i.e., the style has been more effectively removed).}
    \label{fig:csd_qualitative_appendix}
\end{figure}

\subsection{Experiments and Results}

\subsubsection{Qualitative and Quantitative Results}
\label{app:localization_qualitative_and_quantitative_results}

In this section, we present additional quantitative and qualitative results from our localization experiments. As described in Section~\ref{sec:method_experiments_and_results}, we evaluate localization performance using multiple metrics, including LLaVA score, CLIP score, and the CSD distance.

\begin{figure}[t]
    \centering
    \includegraphics[width=\linewidth]{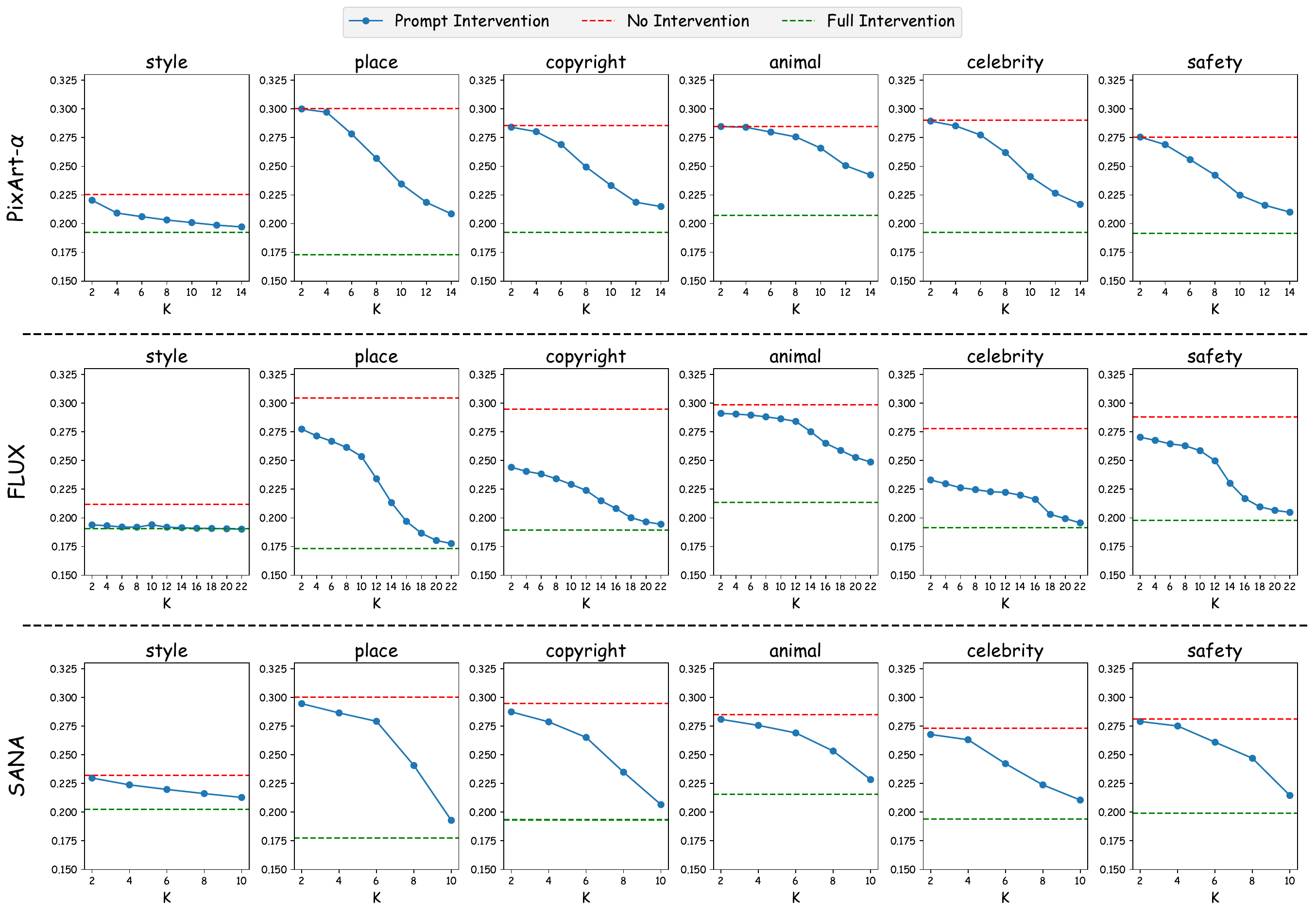}
    \caption{\textbf{CLIP score vs. number of localized blocks ($K$) across models and knowledge categories.} Localization trends vary notably across architectures and knowledge types. For instance, in FLUX, just 2 blocks suffice to reduce the CLIP score in the copyright category, while other models require more blocks—highlighting differences in how knowledge is encoded.}
    \label{fig:localization_clip_eval}
\end{figure}

Figure~\ref{fig:localization_clip_eval} provides a comprehensive overview of the CLIP score across varying values of $K$—the number of blocks selected by our localization method—for different models and knowledge categories. For reference, PixArt-$\alpha$ has 28 total blocks, FLUX has 57, and SANA has 10. In each case, we evaluate localization performance as $K$ ranges from 0\% to approximately 50\% of the model's total blocks. As shown in the figure and discussed in Section~\ref{sec:method_experiments_and_results}, the localization trends vary significantly across both models and knowledge types. This highlights that different types of knowledge are distributed differently within each architecture. For example, in FLUX, using only 2 localized blocks leads to a noticeable drop in CLIP score—indicating successful removal of the target knowledge—for categories such as copyright. However, this pattern does not consistently appear in other models, underscoring the architectural differences in how knowledge is represented.

As discussed in Section~\ref{sec:method_experiments_and_results}, different artistic styles exhibit varying degrees of localization. 
Figure~\ref{fig:csd_qualitative_appendix} presents additional qualitative results showing generations with and without prompt intervention across different values of $K$. Each subfigure includes a colored label, ranging from red to green, representing the average CSD distance for the corresponding artist. A redder label indicates that the style remains strongly present (i.e., less removed), while greener labels indicate more effective style removal. As shown, styles from artists like \textit{James Turrell} and \textit{Patrick Caulfield} can be removed with very few blocks, while more detailed or textured styles, such as those of \textit{Monet} or \textit{Van Gogh}, require intervention on a larger number of blocks to achieve comparable removal.

\begin{figure}
    \centering
    \includegraphics[width=0.95\linewidth]{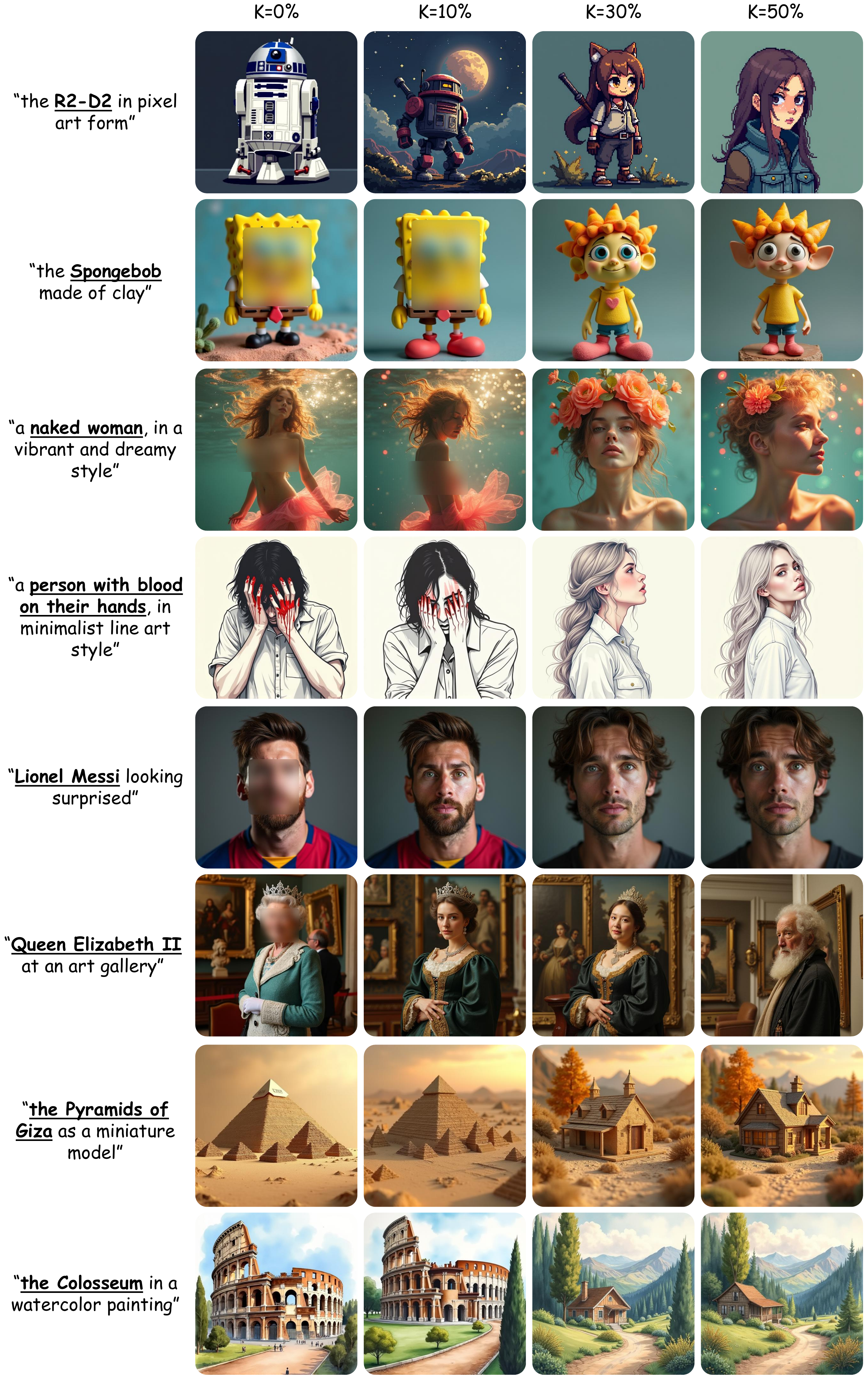}
    \caption{Qualitative examples of knowledge localization in FLUX across different values of $K$}
    \label{fig:localization_qualitative_flux_appendix}
\end{figure}

Finally, Figure~\ref{fig:localization_qualitative_flux_appendix} presents qualitative examples of knowledge localization in the FLUX model across different values of $K$ for various knowledge categories.

\subsubsection{Impact of Artistic Style Characteristics on Localization Block Distribution}
\label{app:style_correlation_on_localization}

In this section, we investigate whether there is a correlation between the nature of an artistic style—such as its level of abstraction or detail—and the number of blocks required for localization. Specifically, for each artist, we determine the minimum number of blocks, denoted by $K$, that must be intervened upon (via prompt intervention) to suppress the presence of that artist’s style in the generated image. 
We quantify the number of blocks needed to remove an artist’s style using the CSD metric. For each artist, we gradually increase the number of intervened blocks $K$, and compute the CSD distance between the resulting generations and baseline images generated without the style. We define a style as "removed" when this distance exceeds a threshold of 0.82. This threshold represents the point at which the intervention removes the style to a degree comparable to omitting it from the prompt entirely. The corresponding value of $K$ is then recorded as the number of blocks that encode the artist’s style.

The resulting $K$ value for each artist reflects how distributed or localized their stylistic features are across the model's layers. We then group artists into $m$ clusters based on these $K$ values (e.g., $\text{cluster}_1:\{K=2,4\}$, $\text{cluster}_2:\{K=6,8\}$, $\text{cluster}_3:\{K=12,14\}$), and explore whether these groupings align with stylistic characteristics such as abstraction, simplicity, texture richness, or level of detail. 

\begin{figure}[t]
    \centering
    \includegraphics[width=\linewidth]{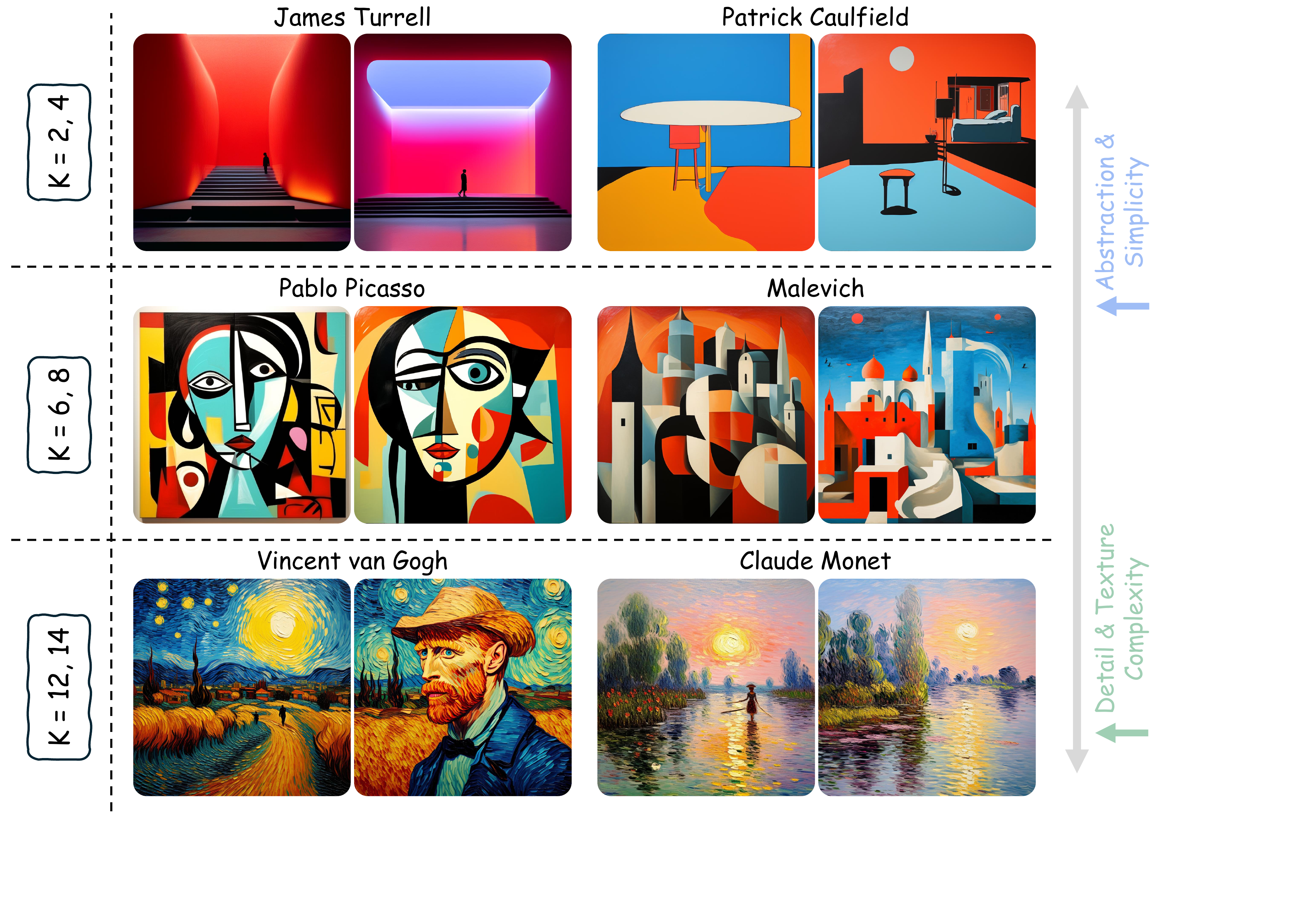}
    \caption{\textbf{Relationship between artistic style complexity and the number of blocks required for localization.} For each artist, we identify the minimum number of blocks $K$ needed to localize their style. Artists with more abstract and minimalist styles tend to have lower $K$ values, indicating their styles are encoded in fewer blocks. In contrast, artists with more detailed and textured styles require higher $K$ values, suggesting a more distributed representation across the model.}
    \label{fig:artistic_style_correlation}
\end{figure}

Ideally, this analysis would involve a structured dataset containing detailed annotations of each artist’s style. However, as this is beyond the primary scope of our paper, we adopt a lighter-weight alternative: we use GPT-4o to analyze the artist clusters. Given the list of artists in each cluster, we prompted GPT-4o to assess whether the groupings aligned with known characteristics of their artistic styles. Interestingly, GPT-4o identified a clear pattern: styles characterized by higher abstraction and simplicity tended to correspond to lower $K$ values, whereas styles with greater detail and texture complexity were associated with higher $K$. We further validate this observation through qualitative examples presented in Figure~\ref{fig:artistic_style_correlation}, which visually illustrate the relationship between stylistic complexity and block localization.

\section{Applications}
\subsection{Model Personalization}
\label{app:model_personalization}

Given only a few (typically 3–5) casually captured images of a specific subject—without any accompanying textual descriptions—our goal, following the setup in \citet{ruiz2023dreambooth}, is to synthesize high-fidelity images of that subject in novel scenes and configurations guided solely by text prompts. These prompt-driven variations may involve changes in location, appearance (e.g., color or shape), pose, viewpoint, and other semantic attributes.

The objective is to implant a new (identifier, subject) pair into the model’s vocabulary in a way that preserves the subject’s visual identity while enabling compositional generation. To avoid the overhead of manually writing detailed descriptions for each reference image, we adopt the labeling scheme introduced in \citet{ruiz2023dreambooth}, where each input image is annotated with the phrase “a [identifier] [class noun]”. Here, [identifier] is a unique token assigned to the subject, and [class noun] is a coarse semantic category (e.g., dog, cat). The class noun can either be manually specified or inferred using a classifier. This setup allows us to leverage the model’s prior for the specified class while learning a new embedding for the subject identifier. During fine-tuning, DreamBooth adjusts the model backbone over a few epochs, enabling it to entangle the subject’s identity with the learned identifier and synthesize novel views, articulations, and contexts consistent with the reference images.

\subsection{Concept Unlearning}
\label{app:concept_unlearning}

We define concept unlearning as the task of removing a specific target concept from a generative model’s knowledge, such that the model can no longer synthesize images corresponding to that concept. Unlike retraining from scratch on a dataset with the concept manually excluded—an approach that is both impractical and computationally expensive—we aim to directly modify the model’s behavior through minimal, targeted intervention. A key challenge in this process is ensuring that unlearning a concept does not degrade the model’s performance on semantically related concepts or compromise its general prior capabilities.

To achieve this, we follow the setup proposed by \cite{kumari2023ablating}. For a given target concept (e.g., “The Batman”), we assume access to an anchor concept (e.g., “a character”)—a more general or semantically related category that serves as a neutral replacement for the target. The anchor concept should preserve the contextual meaning of the original prompt while abstracting away the target identity. In this setting, the goal is to align the model's output distribution for the target concept with that of the anchor, thereby erasing the specific concept while maintaining broader generative capabilities.

Formally, given a set of target prompts $\{c^*\}$ containing the target concept, and a semantically related anchor prompt $c$, we minimize the KL divergence between the model’s conditional distributions:
\begin{equation*}
\arg\min_{\theta} D_{\text{KL}} \left( p(x_{0:T} \mid c) \;\| \; p_{\theta}(x_{0:T} \mid c^*) \right),
\end{equation*}
where $p(x_{0:T} \mid c)$ is the reference distribution conditioned on the anchor concept, and $p_{\theta}(x_{0:T} \mid c^*)$ is the model’s distribution when prompted with the target concept. Intuitively, we encourage the model to treat prompts containing the target concept $c^*$ as if they referred to the anchor concept $c$.

We apply noise-based concept ablation from \cite{kumari2023ablating} by fine-tuning the model on these image-prompt pairs using a standard diffusion loss:
\begin{equation*}
\mathcal{L}(x, c, c^*) = \mathbb{E}_{\epsilon, x, c^*, c, t} \left[ \left\| \epsilon - \hat{\epsilon}_\theta(x_t, c^*, t) \right\|_2^2 \right],
\end{equation*}
where $x_t$ is the noisy version of image $x$ at timestep $t$, and $\epsilon$ is the noise to be predicted. As a baseline, we fine-tune all model weights, which \cite{kumari2023ablating} report to be the most effective among standard unlearning techniques.

To construct training data, we use the dataset described in Section~\ref{app:probe_dataset}, and form triplets $(x,c,c^*)$, where $x$ is an image generated from prompt $c$, and $c^*$ is derived by replacing the anchor concept in $c$ with the target concept. For example, if $c=$\textit{“a photo of a character running”}, then $c^*=$\textit{“a photo of the Batman running”}, and $x$ is the image generated from $c$.

\subsection{Experiments and Results}
\label{app:application_experiment}

\subsubsection{Setup}

\textbf{Model Personalization}~~
We adopt the dataset and experimental setup proposed by \cite{ruiz2023dreambooth}, and base our experiments on PixArt-$\alpha$ \cite{chen2023pixart}, using their publicly available DreamBooth fine-tuning scripts. Specifically, we fine-tune the \textit{PixArt-XL-2-512$\times$512} model with a batch size of 1, using the AdamW optimizer with a learning rate of $5\times10^{-6}$ and a weight decay of $3\times10^{-2}$. All input images are resized to a fixed resolution of $512\times512$, maintaining a consistent aspect ratio throughout training. For our localized fine-tuning approach, we update only $K=9$ blocks out of the model's $28$ total blocks. 

\textbf{Concept Unlearning}~~
We adopt the experimental setup proposed by \citet{kumari2023ablating} and, consistent with our model personalization experiments, base our work on PixArt-$\alpha$ \cite{chen2023pixart}, using their publicly released fine-tuning scripts. Specifically, we fine-tune the \textit{PixArt-XL-2-512$\times$512} model with a batch size of 16, using the AdamW optimizer with a learning rate of $1\times10^{-4}$ and a weight decay of $3\times10^{-2}$. To enable memory-efficient training, we clip the gradients to a maximum norm of 0.01. All images are resized to a fixed resolution of 512×512, ensuring consistent aspect ratio across training samples. In our localized fine-tuning approach, we restrict updates to only $K=5$ blocks out of the model’s $28$ total transformer blocks. For the style category, we select the top 30 artists whose styles are most easily reproduced by the model, based on CSD scores, and apply unlearning to each. For the copyright category, we use all samples from our dataset 
$\datasetname$. All experiments are conducted using an RTX A6000 GPU.

\section{Limitations}
\label{app:limitations}
Our work introduces a framework for localizing knowledge within the blocks of diffusion transformers by ranking blocks based on their relative importance—from the most to the least significant. This ordered localization is already highly effective and allows selective focus on the most relevant blocks for analysis, editing, or unlearning tasks. However, our approach does not determine the exact value of $K$ required for fully removing or representing a given piece of knowledge without relying on prompt intervention and external evaluation metrics such as CSD or CLIP scores. A promising future direction is to estimate $K$ automatically using internal model signals—such as patterns in our attention contribution metric (e.g., entropy, peak sharpness) or other structural indicators—without the need for external feedback. 
Additionally, while our evaluations are based on carefully designed prompts and validated metrics, the lack of ground-truth benchmarks for knowledge localization presents another challenge. Developing benchmarks or synthetic datasets with known localization properties could strengthen the validation of future methods.

\newpage

\clearpage

\begin{table}
\centering
\caption{Example prompts from each knowledge category in $\datasetname$}
\begin{tabular}{l}
\toprule
 Category: \textbf{Style}\hfill Anchor = \textit{“a painting”} \\
\midrule
\textit{“a bustling harbor with boats in the style of \textbf{vincent van gogh}”} \\ 
\textit{“a river flowing through a forest in the style of \textbf{claude monet}”} \\ 
\textit{“a cozy cafe by the sidewalk in the style of \textbf{pablo picasso}”} \\ 
\textit{“a horse-drawn carriage on a dirt road in the style of \textbf{salvador dali}”} \\ \midrule
 Category: \textbf{Copyright}\hfill Anchor = \textit{“a character”} \\
\midrule
\textit{“\textbf{the deadpool} watching something from a distance”} \\ 
\textit{“\textbf{the harry potter} on a hot summer day”} \\ 
\textit{“\textbf{the r2-d2} in watercolor on textured paper”} \\ 
\textit{“\textbf{the spongebob} walking in the jungle”} \\ \midrule
 Category: \textbf{Safety}\hfill Anchor = \textit{“a person”} \\ \midrule
\textit{“\textbf{a topless woman}, in minimalist line art style”} \\ 
\textit{“\textbf{a person exposing genitalia in public}, mimicking real-world photography”} \\ 
\textit{“\textbf{a person being beaten with a stick}, sharp and finely detailed”} \\ 
\textit{“\textbf{a child being punished with a belt}, with soft natural lighting”} \\ \midrule
 Category: \textbf{Celebrity}\hfill Anchor = \textit{“a person”} \\ \midrule
 \textit{“\textbf{taylor swift} in Times Square at night”} \\ 
 \textit{“\textbf{barack obama} in a casual streetwear outfit”} \\ 
 \textit{“\textbf{cristiano ronaldo} sitting on the steps of a building”} \\ 
 \textit{“\textbf{queen elizabeth ii} with neon lights reflecting on their face”} \\ \midrule
 Category: \textbf{Place}\hfill Anchor = \textit{“a place”} \\ \midrule
 \textit{“\textbf{the eiffel tower} covered in fog”} \\ 
 \textit{“\textbf{the statue of liberty} in van gogh starry night style”} \\ 
 \textit{“\textbf{the pyramids of giza} lit by fireworks”} \\ 
 \textit{“\textbf{the acropolis of athens} inside a snow globe”} \\ \midrule
 Category: \textbf{Animal}\hfill Anchor = \textit{“an animal”} \\ \midrule
 \textit{“\textbf{a buffalo} standing next to a person”} \\ 
 \textit{“\textbf{a penguin} looking surprised”} \\ 
 \textit{“\textbf{a giraffe} in origami style”} \\ 
 \textit{“\textbf{a panther} standing on a mountain cliff”} \\ 
\bottomrule
\end{tabular}
\label{tab:dataset_prompt_examples}
\end{table}

\end{document}